\documentclass[11pt]{article}

\usepackage[final]{acl}

\usepackage{times}
\usepackage{latexsym}

\usepackage[T1]{fontenc}

\usepackage[utf8]{inputenc}

\usepackage{microtype}

\usepackage{inconsolata}

\usepackage{graphicx}

\usepackage{algorithm}
\usepackage{algpseudocode}
\usepackage{multirow}
\usepackage{booktabs}
\usepackage{tabularx}
\usepackage{subcaption}
\usepackage{comment}
\usepackage{pifont}
\usepackage{amsmath}
\usepackage{makecell} 
\usepackage{enumitem}

\title{LogicEnvGen: Task-Logic Driven Generation of Diverse Simulated Environments for Embodied AI}

\author{
    Jianan Wang$^{1}$, 
    Siyang Zhang$^{1}$,
    Bin Li$^{2*}$, 
    Juan Chen$^{1*}$, 
    Jingtao Qi$^{2}$, 
    Zhuo Zhang$^{2}$,
    Chen Qian$^{3}$    \\
    $^{1}$College of Computer Science and Technology, National University of Defense Technology \\
    $^{2}$ Intelligent Game and Decision Lab (IGDL), Beijing \\
    $^{3}$ School of Artifical Intelligence, Shanghai Jiao Tong University \\
  \texttt{wangjianan@nudt.edu.cn} ~~~~~ \texttt{libin\_bill@126.com} ~~~~~ \texttt{juanchen@nudt.edu.cn} }

\begin{document}
\maketitle

{\renewcommand{\thefootnote}{}
\footnotetext{*Corresponding authors}}

\begin{abstract}

Simulated environments play an essential role in embodied AI, functionally analogous to test cases in software engineering. 
However, existing environment generation methods often emphasize visual realism (e.g., \textit{object diversity and layout coherence}), overlooking a crucial aspect: logical diversity from the testing perspective.
This limits the comprehensive evaluation of agent adaptability and planning robustness in distinct simulated environments. 
To bridge this gap, we propose \textbf{LogicEnvGen}, a novel method driven by Large Language Models (LLMs) that adopts a top-down paradigm to generate logically diverse simulated environments as test cases for agents.
Given an agent task, LogicEnvGen first analyzes its execution logic to construct decision-tree-structured behavior plans and then synthesizes a set of logical trajectories. 
Subsequently, it adopts a heuristic algorithm to refine the trajectory set, reducing redundant simulation.
For each logical trajectory, which represents a potential task situation, LogicEnvGen correspondingly instantiates a concrete environment.
Notably, it employs constraint solving for physical plausibility.
Furthermore, we introduce LogicEnvEval, a novel benchmark comprising four quantitative metrics for environment evaluation. 
Experimental results verify the lack of logical diversity in baselines and demonstrate that LogicEnvGen achieves 1.08-2.67x greater diversity, significantly improving the performance in revealing agent faults by 3.34\%–72.00\%.

\end{abstract}

\section{Introduction}

\begin{figure}[t]
    \centering
    \includegraphics[width=1.0\columnwidth]{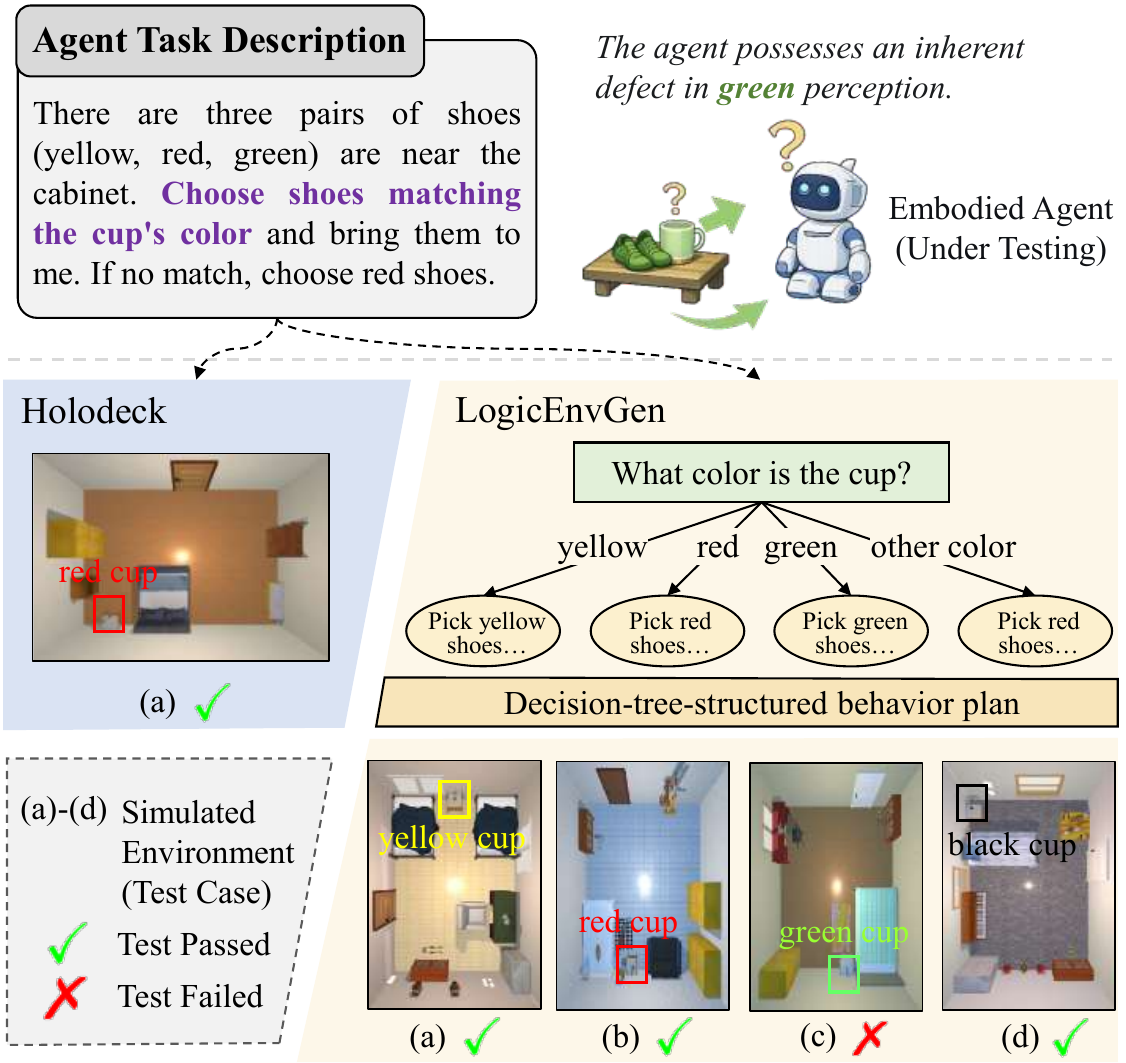}
    \caption{Simulated environments generated by Holodeck \cite{yang2024holodeck} and LogicEnvGen. In contrast, LogicEnvGen generates more logically diverse test cases based on the task description, enabling more comprehensive simulation.}
    \label{fig:figure1}
    \vspace{-12pt}
\end{figure}

Simulation is essential for developing embodied agents, enabling early testing and iterative refinement, thus reducing the risks and costs associated with physical deployment \cite{koenig2004design}.
In this process, simulated environments \cite{DBLP:journals/corr/abs-2503-13882, yang2024holodeck} play a critical role, functioning similarly to test cases in software engineering \cite{tang2024legend}.
The advancement of embodied AI heightens the demand for agents capable of handling complex logical tasks.
This necessitates \textbf{logically diverse} simulated environments that maintain consistent task goals while varying environment states and task conditions, enabling rigorous evaluation of an agent’s adaptability and planning capability across different environment configurations.
Taking the task in Figure \ref{fig:figure1} as an example, creating environments with varied cup colors is necessary for comprehensive testing.

Traditional environment generation methods mainly rely on manual design \cite{kolve2017ai2, deitke2020robothor}, 3D scanning \cite{ramakrishnan2021habitat} and procedural methods \cite{deitke2022️}, which struggle with expensive human effort and limited scalability.
Recently, generative frameworks such as diffusion models are employed to synthesize 3D environments from various inputs \cite{yang2024physcene, tang2024diffuscene}.
With the rise of LLMs, some work utilizes them for environment design, including scene layout and object configurations \cite{wang2023robogen, yang2024holodeck}.
Moreover, some text-to-3D generation methods optimize 3D contents based on pre-trained text-to-image models and focus on localized representations like textures \cite{zhou2024gala3d, he2023t}. 
While these methods present promising capabilities in generating visually realistic simulated environments, they leave logical diversity underexplored, which is pivotal for uncovering agent faults in diverse situations.
In addition, LLM-based methods often face challenges in physical plausibility (e.g., \textit{objects floating in air}).

To bridge this gap, we propose LogicEnvGen, an LLM-driven method that traverses the execution logic of the agent task to generate logically diverse simulated environments as test cases, enabling comprehensive detection of agent faults.  
It adopts a three-phase framework: 
(1) \textit{Behavior Plan Derivation}, to utilize LLMs to decompose the task into independent subtasks and generate a decision-tree-structured behavior plan for each subtask, where each root-to-leaf decision path defines a unique subtask logic flow.
(2) \textit{Logical Trajectory Collection}, to synthesize distinct logical trajectories by combining decision paths from each subtask, where each trajectory represents a potential task situation.
Notably, we employ a heuristic algorithm to prune redundant combinations, thus improving simulation efficiency.
(3) \textit{Simulated Environment Construction}, to leverage LLMs to create an interactive simulated environment for each logical trajectory in a hierarchical and language-guided manner.
To ensure physical plausibility, we formalize object placement as a Constraint Satisfaction Problem (CSP) and incorporate a constraint solver to determine object layouts.
Following a top-down paradigm, LogicEnvGen first captures possible task situations at the abstract logical level and then instantiates them into concrete environments with varied configurations, thus enhancing logical diversity in environment generation.

To assess LogicEnvGen, we introduce \textbf{LogicEnvEval}, a novel benchmark to measure methods that generate embodied environments for use as test cases. 
Unlike existing benchmarks focusing on single-logic and sequential tasks, LogicEnvEval features 50 household tasks with complex execution logic.
Each task is paired with four agent policies (one correct, three faulty).
We also introduce a suite of metrics for environment evaluation: Physics Pass Rate, Logic Coverage, Scenario Validity Rate, and Fault Detection Rate. 
These quantify the environments across three dimensions: physical plausibility, logical diversity, and practical utility as test cases.
Experimental results demonstrate that LogicEnvGen generates physically plausible environments while exhibiting 1.08-2.67x greater Logic Coverage than baselines. 
This significantly improves Fault Detection Rate, ranging from 3.34\% to 72.00\%.
Further evaluations on EMMOE-100 dataset \cite{EMMOE} validate the strong generalization of LogicEnvGen across task distributions.

Our contribution can be summarized as follows.

\setlist[itemize]{itemsep=-3pt, topsep=0pt}

\begin{itemize}

    \item We explore the novel problem of generating logically diverse simulated environments as test cases, improving evaluation of embodied agent adaptability and planning robustness. 

    \item We propose LogicEnvGen, an LLM-driven method that adopts a top-down paradigm to traverse task execution logic for diverse environment generation. It incorporates a heuristic algorithm and constraint solver to reduce redundancy and enhance realism.

    \item We establish LogicEnvEval, a benchmark for evaluating simulated environment generation, along with four quantitative metrics.

    \item Experimental results show LogicEnvGen’s superior performance in physical plausibility, logical diversity, and robust generalization for efficiently revealing agent faults.
    

\end{itemize}   
\section{Related Work}

\subsection{Code Test Generation}
Code testing, a critical part of software development, involves generating various test cases to detect potential bugs and ensure program execution meets expectations.
Existing methods encompass search-based test generation \cite{fraser2011evosuite, lukasczyk2022pynguin} and symbolic test generation \cite{sen2005cute}.  
Recently, LLMs have been employed for automated test generation \cite{schafer2023empirical, ryan2024code}. 
Common evaluation metrics for test generation include accuracy, coverage, and bug detection rate \cite{wang2024testeval, huang2024rethinking}. 
Inspired by this field, we conceptualize simulated environments as test cases and propose a novel method for generating diverse simulated test cases to systematically evaluate embodied agents and reveal potential defects.

\subsection{Embodied AI Environment Generation}
Early studies mainly rely on manual design \cite{li2023behavior}, 3D scanning \cite{deitke2023phone2proc} and procedural generation \cite{fu20213d}, which demand substantial domain expertise and exhibit limited interactivity.
Based on datasets like 3D-FRONT \cite{fu20213d}, some researchers train generative models to generate scenes \cite{tang2024diffuscene, wu2025neural}, which are inherently restricted by the scale and quality of data.
Recently, several studies use LLMs to automatically generate environments \cite{wang2023robogen, feng2023layoutgpt}.
However, they often allow LLMs to directly assign object coordinates, which may violate physical laws like gravity and collision \cite{jia2024cluttergen}.
Some work \cite{yang2024holodeck, littlefair2025flairgpt} formalizes object placement as a constraint satisfaction problem.
Other text-guided 3D scene generation works \cite{zhou2024gala3d, zhou2025layoutdreamer, he2023t} focus on localized representations with limited composability, thus restricting their applicability in embodied AI.
These existing methods emphasize rendering fidelity and visual diversity but overlook logical diversity—a gap that our method is designed to address.


\begin{figure*}[htbp]
    \centering
    \includegraphics[width=\linewidth]{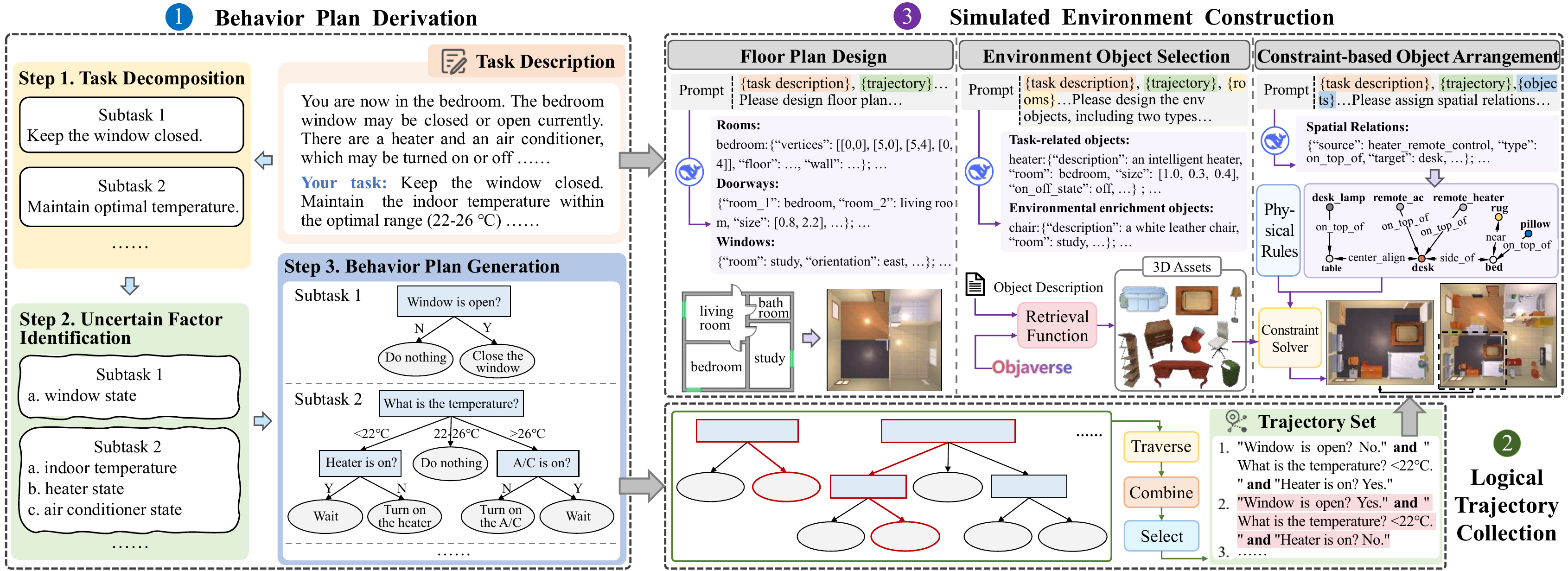}
    \caption{Overview. 
            Phase \ding{172}: Behavior Plan Derivation, decomposes the given task into independent subtasks, identifies uncertain environment factors impacting subtask execution, and generates decision-tree-structured behavior plan for each subtask.
            Phase \ding{173}: Logical Trajectory Collection, traverses and combines decision paths across trees to synthesize distinct logical trajectories for the entire task, each representing a potential task situation.
            Phase \ding{174}: Simulated Environment Construction, instantiates a concrete and physically plausible environment for each situation through three stages: floor plan design, environment object selection and   arrangement.}
    \label{fig:overview}
\end{figure*}

\subsection{LLM for Embodied AI}
With rich world knowledge and remarkable reasoning capabilities, LLMs are extensively employed for Embodied AI \cite{hao2023reasoning}.
Early work leverages LLMs as planners to decompose user instructions into action sequences \cite{ahn2022can, yao2022react}.
Some studies generate executable code to bridge high-level task descriptions and low-level execution \cite{liang2022code}.
Vision-Language-Action (VLA) models integrate perception and reasoning into a unified framework for end-to-end control \cite{VLA-Adapter, PEAfowl}.
In this work, rather than using LLMs as planning engines for embodied agents, we utilize them to construct test cases for simulation.

\section{Problem Formalization}

\textbf{Definition 1} (Logical Diversity). 
Consider a task $T$ defined over a state space $\mathcal{S}$, where each state $s\in \mathcal{S}$ is defined by a set of variables $\mathcal{V}$. 
Let $\mathcal{V}_{L} \subseteq \mathcal{V}$ denote the subset of variables that govern the execution logic of $T$, where each variable $v \in \mathcal{V}_{L}$ has a finite discrete domain $\mathcal{D}_v$.
The set of all logically possible atomic conditions of $T$ can be represented as $\mathcal{C}_{\text{total}} = \{(v, d) \mid v \in \mathcal{V}_{L}, d \in \mathcal{D}_v\}$.

Given a set of simulated environments $\mathcal{E}$, we define $\mathcal{C}_{\mathcal{E}} \subseteq \mathcal{C}_{\text{total}}$ as the set of conditions covered by $\mathcal{E}$. 
Specifically, a condition $(v, d)$ is in $\mathcal{C}_{\mathcal{E}}$ if there exists at least one environment $e \in \mathcal{E}$ where variable $v$ takes the value $d$.
The logical diversity of $\mathcal{E}$ is quantified as the coverage ratio:
\begin{equation}
    \label{eq:logical_diversity}
    \text{LD}(\mathcal{E}) = \frac{|\mathcal{C}_{\mathcal{E}}|}{|\mathcal{C}_{\text{total}}|}
\end{equation}

\section{Method}

\subsection{Overview}
In this paper, we propose LogicEnvGen, an LLM-driven method to generate logically diverse test cases for the rigorous detection of agent deficiencies.
Adopting a top-down scheme, LogicEnvGen captures all potential situations at the abstract logical level and instantiates concrete environments, enabling coverage of diverse task decision paths.
As shown in Figure \ref{fig:overview}, it comprises three phases. 

\subsection{Behavior Plan Derivation}
Given an agent task, we utilize LLMs to analyze its execution logic to derive decision-tree-structured behavior plans.
Specifically, the process includes three steps: task decomposition, uncertain factor identification and behavior plan generation.

In contrast to prior work that primarily divides tasks into sequentially executed subtasks, we decompose the embodied task into mutually independent subtasks without execution order dependencies.
This effectively avoids capturing logically redundant trajectories.
For each subtask, we identify the associated uncertain environmental factors affecting the agent execution logic.    
Subsequently, we instruct LLMs to construct a decision-tree-structured, environment-aware task plan based on these factors, which outlines distinct agent behavior logic under diverse conditions. 
Represented in JSON format, its internal nodes are conditional \textit{queries} about the environment (e.g., ``\textit{Is the window open?}'') and the edges correspond to the \textit{responses} (e.g., ``\textit{Yes}'').
The root-to-leaf decision path defines a unique logic flow, representing a potential situation for the subtask.
The leaf node encapsulates the agent execution plan in the corresponding situation.

To ensure the reliability and logical consistency of the derived plans, we employ a rule-guided, iterative verification and refinement process. It checks for: 
(a) the mutual independence of subtasks by detecting overlapping uncertain factors, 
(b) the syntactic correctness of the generated tree-structured plans, and 
(c) the grounding of the internal tree nodes in the identified factors.

\begin{algorithm}[hbp]
    \caption{Minimal Trajectory Selection}
    \label{algorithm}
    \textbf{Input}: Trajectory set $\mathcal{T}$      \\
    \textbf{Output}: Minimal trajectory set $\mathcal{T}_{min}$
    
    \begin{algorithmic}[1]
    
        \State $\mathcal{T}_{min}, \mathcal{C}_{pool}, \mathcal{T}_{cand} \gets \{\}, \{\}, \{\}$ \Comment{initialize}
        
        \For{ $t \in \mathcal{T}$}
            \State $\mathcal{C}_t \gets \text{splitConstraints}(t)$ 
            \If {$\mathcal{C}_t \cap \mathcal{C}_{pool} = \emptyset$}  
                \State $\mathcal{T}_{min} \gets \mathcal{T}_{min} \cup \{t\}$ \Comment{prioritize}
                \State $\mathcal{C}_{pool} \gets \mathcal{C}_{pool} \cup \mathcal{C}_t$ \Comment{update}
            \ElsIf{$\mathcal{C}_t \not\subseteq \mathcal{C}_{pool}$} 
                \State $\mathcal{T}_{cand} \gets \mathcal{T}_{cand} \cup \{t\}$ \Comment{candidate}
            \EndIf
        \EndFor
        \State $\mathcal{T}_{min} \gets \text{Update}(\mathcal{T}_{min}, \mathcal{T}_{cand}, \mathcal{C}_{pool})$ \Comment{check}
        \State \Return $\mathcal{T}_{min}$
    \end{algorithmic}
\end{algorithm}

\subsection{Logical Trajectory Collection}
Based on the tree-structured plan of each subtask, we use depth-first search (DFS) to extract the set of decision \textit{paths}.
Each path, comprising multiple query-response pairs, is considered as a \textit{constraint}.
The path sets from all subtasks are subsequently integrated to form a set of logical \textit{trajectories} for the entire task.
Each trajectory, consisting of several constraints, represents a possible situation that agent may encounter.
A direct integration approach is via Cartesian product computation to enumerate all trajectories. 
However, it can cause a combinatorial explosion as trajectories scale exponentially with the number of subtasks, necessitating computationally expensive simulations under excessive environments.
Furthermore, most trajectories fail to introduce new constraints, making them redundant in covering logical diversity.

\begin{figure}[tp]
    \centering
    \includegraphics[width=1.0\linewidth]{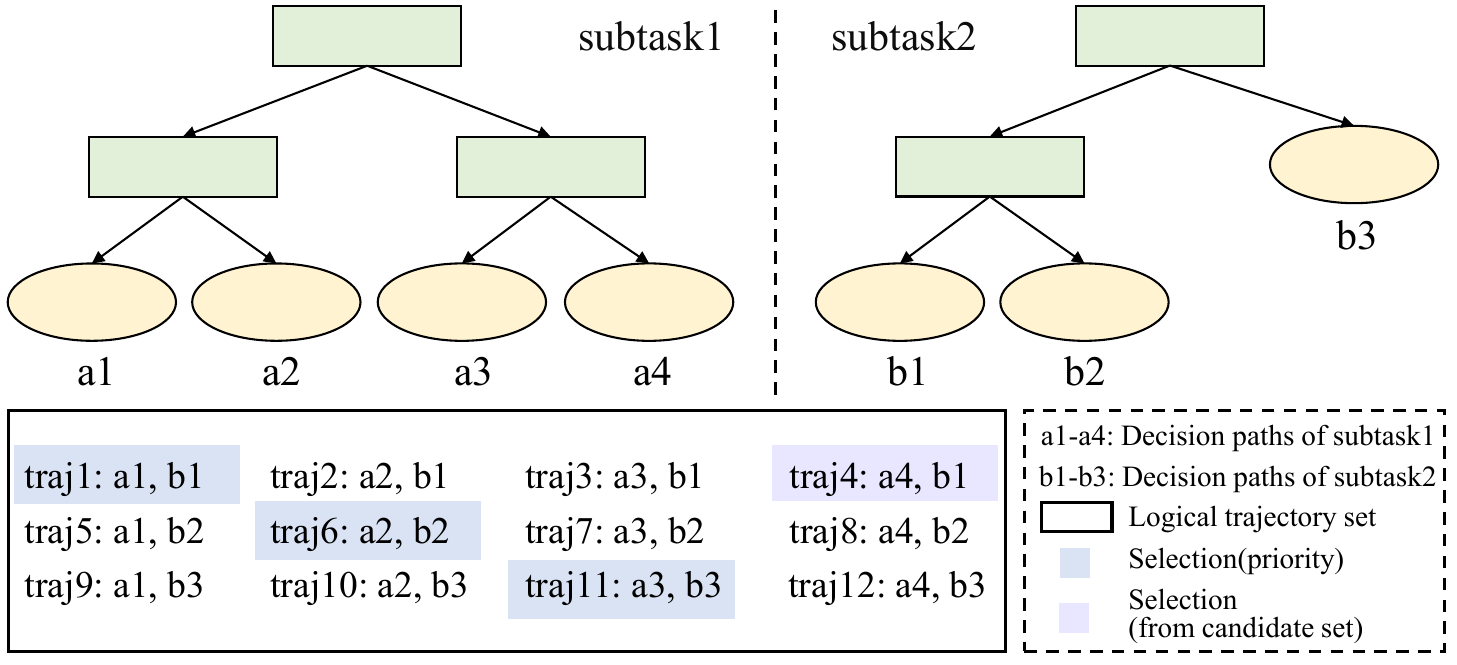}
    \caption{Algorithm example. Logical trajectories reduce from 12 to 4 (3 blue, 1 purple).}
    \label{fig:algorithm-example}
\end{figure}

To address this, we employ a heuristic Minimal Trajectory Selection algorithm to eliminate redundant trajectories while preserving a minimal subset covering all constraints, as outlined in Algorithm \ref{algorithm}.
Specifically, we initialize an empty constraint pool.
Based on the logical trajectory set derived from Cartesian product, we prioritize selecting trajectories whose constituent constraints are entirely disjoint from the current pool, while extracting those exhibiting partial constraint non-overlap into a candidate set.
After each selection, its constituent constraints are added to the pool.
Finally, we check the candidate set to select remaining trajectories whose constraints have not yet been integrated into the pool.
An example is shown in Figure \ref{fig:algorithm-example}.

By constructing the tree-structured behavior plans and collecting all logical trajectories, we achieve a comprehensive capture of potential task situations.
This enables the logical diversity of environments to be controlled at an abstract level.
Furthermore, the employed algorithm ensures diversity coverage with a minimal set of situations, reducing redundant simulation and shortening the testing cycle. 
The ablation and comparative results (Sections \ref{sec:ablation-1} and \ref{sec:ablation-2}) validate the efficacy of the tree-structured plans and the algorithm.

\subsection{Simulated Environment Construction}
For each logical trajectory in the minimal set, we create a text-based test case and instantiate it in AI2-THOR \cite{kolve2017ai2}.
Leveraging the scene planning capability of LLMs, this process is language-guided and can be deconstructed into three stages:
(a) \textit{Floor Plan Design}, determining room layouts and placements of doors and windows; 
(b) \textit{Environment Object Selection}, specifying the environment objects and retrieving their corresponding 3D assets;
(c) \textit{Constraint-based Object Arrangement}, organizing objects based on designed spatial relations and real-world physical rules.

\subsubsection{Floor Plan Design} 
The floor plan defines the foundational structure of the simulated environment, comprising three components: rooms, doorways, and windows.
Specifically, each room is delineated as a rectangle by its vertex coordinates and is characterized by its floor/wall colors and materials.
Doorways are specified by their types, sizes, and states (e.g., \textit{opened}, \textit{closed}).
And windows are determined by attributes like orientations (e.g., \textit{west}, \textit{east}), types, and states.

\subsubsection{Environment Object Selection} 
Objects serve as spatial infill and are grouped into task-related and environmental enrichment types.
This grouping allows the environment to be tailored to specific task requirements and trajectory constraints, while being enriched by integrating contextually appropriate objects.
Each object is defined by properties like a textual description, assigned room, size, and relevant attributes (e.g., \textit{on\_off\_state}).
Based on the textual descriptions (e.g., \textit{``a brown two-seater sofa''}), we retrieve the corresponding 3D assets from Objaverse \cite{deitke2023objaverse}.
This retrieval process is automated through CLIP \cite{radford2021learning} to match descriptions with assets via text-image similarity scores, following prior work such as Holodeck \cite{yang2024holodeck}.

\subsubsection{Constraint-based Object Arrangement}
Due to the inherent limitations of LLMs in numerical reasoning, directly generating precise spatial coordinates and orientations for multiple objects often violates task requirements and physical principles.
To circumvent this, we utilize a hybrid approach wherein LLMs first define high-level spatial relation constraints, after which a computational constraint solver determines the final object arrangements.
This ensures that object placements align with task-specific requirements and trajectory constraints, while maintaining a coherent environment structure and adherence to physical laws.

We define four categories of spatial relations:
(1) Unary: edge, center, mounted on wall;
(2) Contact: in, on top of;
(3) Distance: near, far;
(4) Relative: above, in front of, side of, center aligned, face to.
However, LLM-generated relations often show logical inconsistencies (e.g., \textit{``pen is in drawer''} and \textit{``pen is on top of bed''}).
To address such conflicts, we employ a rule-based physical compatibility checker to evaluate their plausibility.
Detected inconsistencies trigger iterative refinement, prompting the LLM to revise its output based on feedback.

Subsequently, the object arrangement task is modeled as a constraint satisfaction problem (CSP), using the Z3 solver\footnote{\url{https://github.com/Z3Prover/z3}} to compute solutions.
Here, object positions and directions (alongside doorway and window positions) constitute variables. 
Spatial relations and physical laws (e.g., \textit{non-floating-in-air}, \textit{non-colliding}) act as constraints.
Notably, if no solution satisfies all constraints, we adopt a constraint relaxation mechanism. 
This selectively disregards peripheral spatial relations, particularly those involving environmental enrichment objects, to prioritize essential task-related constraints.
Section \ref{sec:ablation-3} shows the efficacy of Z3 via ablation.

\section{Experiments}
\label{sec:experiments}

\subsection{Benchmarks}
To comprehensively evaluate the proposed LogicEnvGen, we conduct experiments on a novel benchmark, LogicEnvEval, and further validate its generalization on EMMOE-100 \cite{EMMOE}.

\begin{figure}[tp]
    \centering
    \includegraphics[width=\linewidth]{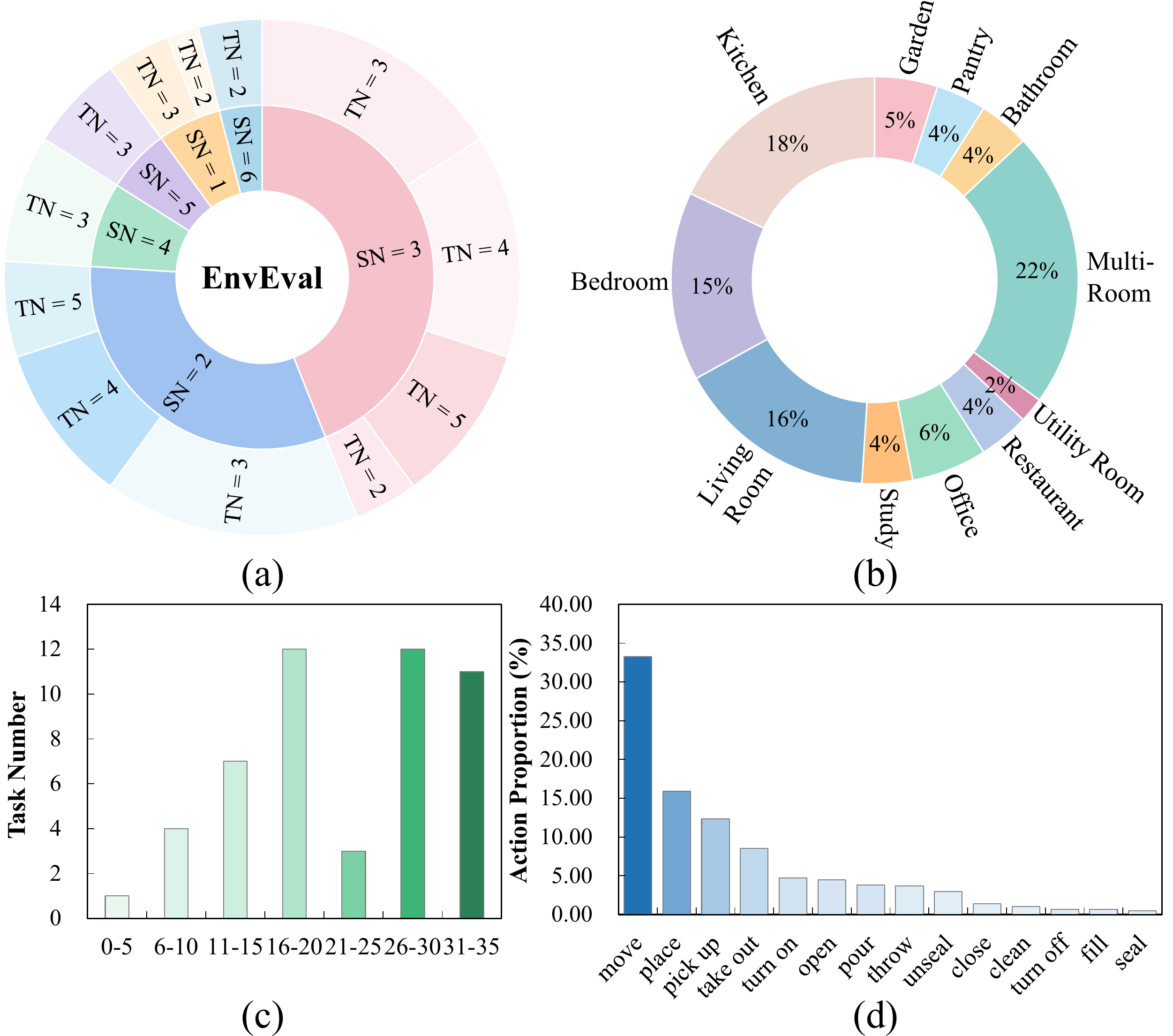}
    \caption{LogicEnvEval Benchmark. (a) Task composition. SN: number of subtasks. TN: number of task trajectories (minimal). (b) Environment type distribution of all subtasks. (c) Action step distribution of correct policies. (d) Action type diversity in correct policies.}
    \label{fig:benchmark}
    \vspace{-10pt}
\end{figure}

\paragraph{LogicEnvEval:}
Based on Bode et al.'s definition of \textit{conditional tasks} \cite{bode2024comparison}—logically complex tasks requiring agents to gather environment information and adapt dynamically—we construct LogicEnvEval, a dataset comprising 50 long-horizon complex tasks. 
Figure \ref{fig:benchmark} shows its composition and the distribution of environment types, task steps, and agent action types.
Specifically, these tasks involve household activities, and each is paired with a ground-truth decision-tree-structured behavior plan. 
Given the prevalence of behavior trees (BTs) in robotics \cite{colledanchise2018behavior}, we adopt them to represent agent policies. 
For each task, we construct one correct agent policy and three faulty policies based on Code-BT \cite{zhangcode}. 
Building upon existing taxonomies of faulty policies \cite{wang2025besimulator}, we categorize the faulty BTs into three distinct types: Counterfactuals, Unreachable, and Lackbranch. 
See Appendix \ref{appendix:EnvEval} for more LogicEnvEval information.

\paragraph{EMMOE-100:}
It features 100 everyday household tasks with various attributes, rigorously challenging the logical reasoning and planning capabilities of agents in continuous environments.

\subsection{Experiment Setup}

\subsubsection{Metrics}
Following the evaluation criteria in code test generation \cite{ryan2024code}, we introduce four novel metrics for simulated environment generation.

\begin{table}[htp]
    \centering
    \small
    
    \begin{tabularx}{\linewidth}{lX}
        \toprule
            \textbf{Dimension}   & \textbf{Physical Constraints}   \\
        \midrule
            \multirow{5}{*}{\centering Floor Plan} 
                & Adjacent rooms must not overlap.   \\ [0.5ex]
                \cline{2-2} 
                \addlinespace[0.5ex]
                & Doors and windows must be embedded in the walls, with doors placed on the floor and windows positioned above it.  \\
        \midrule
            \multirow{2}{*}{Entity}   & Objects must not collide with one another and must remain within room boundaries.\\
        \midrule
            \multirow{2}{*}{Relation} & The positions and directions of objects must satisfy the specified spatial relations.  \\
        \bottomrule            
    \end{tabularx}

    \caption{\label{tab:Physics Pass Rate}
    Three dimensions of Physics Pass Rate.}
    \vspace{-10pt}
\end{table}

\begin{table*}[htbp]
    \centering
    \fontsize{8.1}{11}\selectfont
    \setlength{\tabcolsep}{2.1pt}

    \begin{tabular*}{\textwidth}{l l cccc c c cccc c}
        \toprule
            \multirow{2}{*}{\textbf{Model}} & \multirow{2}{*}{\textbf{Method}} & \multicolumn{4}{c}{\textbf{Physics Pass Rate (\%)}} & \multicolumn{1}{c}{\multirow{2}{*}{\textbf{LogCov}}} & \multicolumn{1}{c}{\multirow{2}{*}{\textbf{SceVR}}} & \multicolumn{4}{c}{\textbf{Fault Detection Rate (\%)}} & \multicolumn{1}{c}{\multirow{2}{*}{\textbf{ST}}} \\
            
            \cmidrule(lr){3-6}
            \cmidrule(lr){9-12}
        
            & & \textbf{Floor Plan} & \textbf{Entity} & \textbf{Relation} & \textbf{Avg} & \multicolumn{1}{c}{\textbf{(\%)}} & \multicolumn{1}{c}{\textbf{(\%)}} & \textbf{CFactuals} & \textbf{Unreach} & \textbf{LBranch} & \textbf{Avg} & \multicolumn{1}{c}{\textbf{(min)}} \\

            \midrule
            \multirow{5}{*}{\textbf{DeepSeek-v3.2}}  & CoT        & 98.00      & 26.66  & 19.66    & 48.11   
                                                       & 73.08                                            
                                                       & 84.48                                           
                                                       & 70.00      & 64.00  & 64.00    & 66.00   
                                                       & 35.86 \\
                                          & IFG        & 89.82      & 11.16  & 7.70    & 36.23
                                                       & 91.78
                                                       & 90.25
                                                       & 90.00      & \textbf{90.00}  & 88.00    & 89.33  
                                                       & 98.21 \\
                                          & Holodeck   & 100.00     & 100.00 & 100.00   & \textbf{100.00}
                                                       & 37.07                                            
                                                       & 74.00                                            
                                                       & 24.00      & 22.00  & 28.00    & 24.67   
                                                       & 14.71 \\
                                          & I-Design   & -          & 86.00  & 18.00   & 52.00
                                                       & 37.07                                            
                                                       & 64.00                                            
                                                       & 20.00      & 18.00  & 24.00    & 20.67   
                                                       & 13.93 \\
                                          & \textbf{LogicEnvGen}    & 100.00     & 100.00 & 100.00   & \textbf{100.00}
                                                       & \textbf{98.97}                                         
                                                       & \textbf{92.40}                                            
                                                       & \textbf{94.00}      & \textbf{90.00}  & \textbf{94.00}   & \textbf{92.67}  
                                                       & 51.42 \\
            \midrule
            \multirow{5}{*}{\textbf{Gemini-2.5-Flash}} & CoT   & 91.23      & 4.00   & 12.59     & 35.94   
                                                       & 89.50                                            
                                                       & 81.34                                             
                                                       & 84.00      & 88.00  & 72.00    & 81.33         
                                                       & 54.64 \\
                                          & IFG        & 72.50      & 12.32  & 23.90    & 36.24
                                                       & 95.40
                                                       & 88.36
                                                       & \textbf{92.00}     & 90.00    & 88.00   & 90.00 
                                                       & 88.49 \\
                                          & Holodeck   & 100.00     & 100.00 & 100.00   & \textbf{100.00}  
                                                       & 37.07                                            
                                                       & 68.00                                            
                                                       & 26.00      & 22.00  & 22.00    & 23.33   
                                                       & 16.26 \\
                                          & I-Design   & -          & 92.00  & 14.00   & 53.00
                                                       & 37.07                                            
                                                       & 70.00                                            
                                                       & 28.00      & 24.00  & 24.00    & 25.33   
                                                       & 14.62 \\
                                          & \textbf{LogicEnvGen}    & 100.00     & 100.00 & 100.00   & \textbf{100.00}
                                                       & \textbf{97.29}                                      
                                                       & \textbf{91.83}                                      
                                                       & \textbf{92.00}      & \textbf{92.00} & \textbf{90.00} & \textbf{91.33} 
                                                       & 52.82 \\
            \midrule
            \multirow{5}{*}{\textbf{Qwen2.5-72B}}  & CoT        & 86.00      & 24.00  & 25.00    & 45.00   
                                                       & 66.16                                            
                                                       & 73.99                                            
                                                       & 58.00      & 60.00  & 58.00    & 58.67   
                                                       & 35.16 \\
                                          & IFG        & 88.58      & 12.78  & 22.46    & 41.27
                                                       & 93.01
                                                       & 73.69
                                                       & 82.00      & 76.00  & \textbf{90.00}    & 82.67   
                                                       & 95.57 \\
                                          & Holodeck   & 100.00     & 100.00 & 100.00   & \textbf{100.00}  
                                                       & 37.07                                            
                                                       & 60.00                                            
                                                       & 16.00      & 16.00  & 14.00    & 15.33   
                                                       & 14.19 \\
                                          & I-Design   & -          & 82.00  & 10.00   & 46.00
                                                       & 37.07                                            
                                                       & 56.00                                            
                                                       & 18.00      & 14.00  & 12.00    & 14.67   
                                                       & 14.07 \\
                                          & \textbf{LogicEnvGen}       & 100.00     & 100.00 & 100.00   & \textbf{100.00}  
                                                       & \textbf{93.54}                                      
                                                       & \textbf{82.40}                             
                                                       & \textbf{86.00}      & \textbf{84.00}  & 88.00    & \textbf{86.00}   
                                                       & 55.12 \\
            \bottomrule
    \end{tabular*}

    \caption{\label{tab:experiment performance}
    The comparative experiment results on the LogicEnvEval benchmark.}
\end{table*}

\begin{itemize}
    \item \textbf{Physics Pass Rate (PhyPR)} 
    This metric assesses the adherence of environments to physical constraints in three dimensions (Table \ref{tab:Physics Pass Rate}). 
    
    \item \textbf{Logic Coverage (LogCov)} 
    This metric measures the extent to which the environments cover the decision paths of the ground-truth decision trees.
    It indirectly quantifies logical diversity as a hierarchical composition of atomic conditions.
    For automatic evaluation, we develop a rule-based script tool to parse environment metadata to identify the paths covered by an environment.
    
    \item \textbf{Scenario Validity Rate (SceVR)}
    This metric represents the proportion of valid environments where correct agent policies can be simulated and verified as fault-free.
    Specifically, an environment is invalid if it lacks task-related objects, violates explicit task constraints (e.g., \textit{a coat is initialized on the bed, but the task requires it to be in the closet}), etc.
    
    \item \textbf{Fault Detection Rate (FauDR)} 
    This metric measures the fault detection efficacy of the generated simulated environments. 
    Based on 150 faulty BTs of LogicEnvEval, we conduct BT simulation in the generated environments and compute the proportion of the revealed faulty BTs. 
    To ensure fairness, we exclude invalid test environments and only assess the capability of valid ones to reveal genuine faults.
    This is because invalid environments invariably report faults due to their configuration errors, which is not effective fault detection.
    
\end{itemize}

For BT simulation, we employ a behavior simulator \cite{wang2025besimulator} to efficiently execute BTs and validate whether they are faulty.
Besides, we use Simulation Time (ST) to assess the time to verify a BT on the generated test environments.

\subsubsection{Baselines and Adopted LLMs}
Given our focus on generating logically diverse simulated environments as test cases, we compare LogicEnvGen against four baselines: 
the environment generation method Holodeck \cite{yang2024holodeck}, I-Design \cite{I-Design}, 
the general LLM reasoning method CoT \cite{wei2022chain}, 
and IFG \cite{ahmed2025intent}, a method designed to boost LLM semantic diversity.
The CoT and IFG baselines are adapted for diverse environment generation.
More details are in Appendix \ref{appendix:Baseline}.
For fairness, all methods use the same problem setup and 3D asset library. 
To demonstrate the cross-LLM generalization of LogicEnvGen, we use three well-known LLMs: DeepSeek-v3.2 \cite{deepseek-v3}, Gemini-2.5-Flash \cite{gemini-2.5-flash} and Qwen2.5-72B-Instruct \cite{yang2024qwen2}.


\subsection{Method Performance on LogicEnvEval}
\label{sec:performance}
Table \ref{tab:experiment performance} presents the performance comparison between LogicEnvGen and baselines.
The results show that LogicEnvGen generates physically plausible simulated environments with broader logical coverage and higher validity rate, and achieves superior fault detection with great efficiency.

In terms of PhyPR, both Holodeck and LogicEnvGen generate simulated environments adhering to physical constraints.
In contrast, CoT and IFG often fail to satisfy physical plausibility, particularly in the Entity and Relation dimensions.  
I-Design struggles to find reasonable placements for large object counts, and the renders may result in visual distortions because of the asset resizing process.


\begin{table}[tb]
    \centering
    \fontsize{8.1}{10}\selectfont
    \setlength{\tabcolsep}{4.5pt}

    \begin{tabular*}{\linewidth}{l ccccc}
    
        \toprule
            \multirow{2}{*}{\textbf{Method}}      & \textbf{PhyPR}      & \textbf{LogCov}      & \textbf{SceVR}   
                                 & \textbf{FauDR}      & \textbf{ST}    \\
                        & \textbf{(\%)}  & \textbf{(\%)} & \textbf{(\%)}  & \textbf{(\%)}  & \textbf{(min)} \\
        \midrule
            CoT                           & 45.26    & 96.67    & 86.67    & 84.33    & 12.29 \\
            IFG                           & 34.29    & 98.33    & 93.43    & 95.00    & 19.28 \\
            Holodeck                      & \textbf{100.00}   & 92.00    & 79.00    & 72.67    & 9.37 \\
            I-Design                      & 49.50    & 92.00    & 66.00    & 60.33    & 8.74 \\
            \textbf{LogicEnvGen}  & \textbf{100.00}  & \textbf{99.67}   & \textbf{94.62}   & \textbf{95.67}   & 11.78 \\
        \bottomrule

    \end{tabular*}

    \caption{The comparative experiment results on the EMMOE-100 benchmark.}
    \label{tab:Emmoe}
    \vspace{-12pt}
\end{table}

Notably, LogicEnvGen offers benefits in two key aspects.
First, incorporating constrained trajectory prompts enables finer-grained guidance for LLMs to construct simulated environments, reducing uncertainty and hallucinations. 
This enhances the SceVR of LogicEnvGen, achieving 3.47\% to 23.83\% improvements over baselines (Gemini-2.5-Flash).
Our analysis reveals that the environments generated by baselines often exhibit critical deficiencies, such as missing task-related objects or incorrect attribute specifications. 
Second, LogicEnvGen generates simulation test cases with 1.08-2.67 times the LogCov of the four baselines (DeepSeek-v3.2).
This demonstrates that constructing tree-structured behavior plans effectively promotes logically diverse test environment generation, facilitating the identification of potential faults in agent policies. 
For instance, LogicEnvGen achieves an average FauDR of 92.67\% across three types of faulty BTs, outperforming baselines by 3.34\% to 72.00\% (DeepSeek-v3.2).
Besides, IFG shows commendable performance in LogCov and FauDR, but its efficiency is significantly hindered by logical redundancy in generated cases, requiring 1.73 times more simulation time on average per BT than LogicEnvGen (Qwen2.5-72B).
Visual quality comparisons on LogicEnvEval are in Appendix \ref{appendix:VisualEvaluation}.

\subsection{Method Performance on EMMOE-100}
To validate the generalization of LogicEnvGen across diverse task and scenario distributions, we evaluate it on EMMOE-100 \cite{EMMOE} using DeepSeek-v3.2. 
Table \ref{tab:Emmoe} shows its superiority across various metrics over the baselines. 
Since EMMOE-100 primarily comprises single-logic and sequential tasks (e.g., \textit{put the apple on the table to the sofa}), baselines such as Holodeck already achieve a high 92.00\% LogCov. 
Furthermore, in agent fault identification, LogicEnvGen slightly outperforms the strongest baseline IFG by 0.67\% while enhancing simulation efficiency by 38.90\%.

\subsection{Ablation Study}
\label{sec:ablation}

We conduct ablation studies on DeepSeek-v3.2 to demonstrate the critical roles of the decision-tree-structured behavior plans (DBP), Minimal Trajectory Selection algorithm (MTSA) and constraint-based arrangement solving (CAS) in enhancing test environment generation.
We use a new evaluation metric, Jaccard Index (JI), to quantify the similarity between the generated logical trajectory set and the minimal ground-truth set. 
JI ranges from 0 to 1, with higher values indicating greater similarity.

\begin{table}[tb]
    \centering
    \fontsize{8.0}{10}\selectfont
    \setlength{\tabcolsep}{0.4pt}

    \begin{tabular*}{\linewidth}{l ccccc}
    
        \toprule
            \textbf{Method}      & \textbf{PhyPR(\%)}      & \textbf{LogCov(\%)}      & \textbf{SceVR(\%)}   
                                 & \textbf{FauDR(\%)}      & \textbf{JI}    \\
        \midrule
            W/O DBP                       & 100.00    & 93.25    & 89.97    & 90.67    & 0.33 \\
            \textbf{LogicEnvGen}          & 100.00    & 98.97    & 92.40    & 92.67    & 0.88 \\
        \bottomrule

    \end{tabular*}

    \caption{Results of ablation study on the decision-tree-structured behavior plan (DBP).}
    \label{tab:ablation_dt}
\end{table}


\begin{table}[tb]
    \centering
    \fontsize{8.0}{10}\selectfont
    \setlength{\tabcolsep}{1.9pt}

    \begin{tabular*}{\linewidth}{l ccccc}

        \toprule
            \textbf{Method}  & \textbf{LogCov(\%)} & \textbf{JI} & \textbf{TC} & \textbf{SceVR(\%)} & \textbf{FauDR(\%)}\\
        \midrule
            Exhaustive            & 98.97            & 0.92            & O($2^N$)                 & 92.59            & 92.00      \\ 
        \midrule             
            W/O MTSA              & 98.97            & 0.27            & -                        & 91.78            & 94.00      \\  
        \midrule
            \textbf{LogicEnvGen}  & 98.97            & 0.88            & O(N)                     & 92.40            & 92.67      \\  
        \bottomrule
        
    \end{tabular*}
    
    \caption{Results of comparative and ablation study on the Minimal Trajectory Selection Algorithm (MTSA). TC denotes Time Complexity. \textit{N} denotes the number of trajectories generated by the Cartesian product.}
    \label{tab:ablation_algorithm} 
    
    \vspace{-7pt} 
    
\end{table}

\subsubsection{Effectiveness of DBP}
\label{sec:ablation-1}
We ablate DBP by removing decision tree generation and guiding LLMs to synthesize logical trajectories by analyzing task logic.
Table \ref{tab:ablation_dt} shows that the LogCov and FauDR decrease by 5.72\% and 2.00\%, respectively.
This highlights that the decision-tree-structured behavior plan explicitly organizes task logic to cover potential situations, facilitating logically diverse environment generation.

\subsubsection{Effectiveness of MTSA}
\label{sec:ablation-2}
We ablate MTSA by directly utilizing the full logical trajectory set derived from the Cartesian product for environment construction. 
Table \ref{tab:ablation_algorithm} reveals a significant decrease of 0.61 in JI, demonstrating the efficacy of our algorithm in eliminating redundant trajectories to enhance simulation efficiency. 
Notably, it maintains a highly comparable FauDR with a marginal difference of 1.33\%.
Furthermore, we compare our algorithm with the Exhaustive algorithm. 
The exhaustive algorithm theoretically guarantees a minimum set-cover, however, it incurs exponential time complexity. 
In contrast, our algorithm exhibits superior computational efficiency.

\begin{figure}[tbp]
    \centering
    \includegraphics[width=\linewidth]{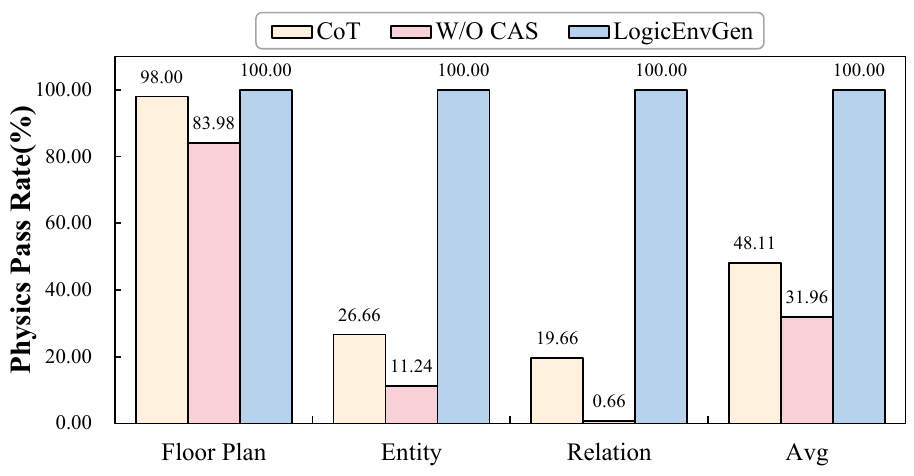}
    \caption{Results of ablation study on the constraint-based arrangement solving (CAS).}
    \label{fig:ablation_z3}

    \vspace{-8pt}
\end{figure}

\subsubsection{Effectiveness of CAS}
\label{sec:ablation-3}
We ablate CAS by removing the Z3 constraint solver and instructing LLMs to design asset placements based on physical laws and spatial relations. 
Figure \ref{fig:ablation_z3} indicates a significant decline in PhyPR across three dimensions. 
Interestingly, the ablation variant underperforms CoT, averaging 16.15\% lower in PhyPR. 
This is because LogicEnvGen designs richer objects and spatial relations than CoT, making it more challenging to construct physically plausible environments. 
These findings highlight LLMs numerical reasoning challenges and validate CAS, which improves PhyPR by 68.04\%.

\section{Conclusion}

We propose LogicEnvGen, an LLM-driven framework for the automated generation of logically diverse simulated environments.
Adopting a top-down paradigm, we first capture all possible task situations at the abstract logical level and then instantiate concrete environment cases, thus yielding broad coverage of the agent task decision paths.
Based on the proposed LogicEnvEval and the existing EMMOE-100 benchmark, we conduct quantitative evaluations across four metrics. 
The results demonstrate that LogicEnvGen exhibits strong generalization, ensuring physical plausibility while achieving significant improvements in logical diversity and fault detection effectiveness.

\section{Limitations}

We employ a behavior simulator to conduct BT simulation to measure the scenario validity rate and fault detection rate of the generated environments. 
We use this simulator for its high efficiency and reliability. 
However, using general physics simulators like Gazebo or Unreal Engine can yield more high-fidelity evaluation results. 
This highlights the need for future evaluation advancement.

Although we utilize a rule-guided iterative verification process to ensure the reliability of the generated behavior plans, relying on LLMs for automated generation remains susceptible to logical omissions and misinterpretations.
The assumption that LLMs can capture complete task execution logic from task descriptions may become fragile for highly complex tasks.
Future work can introduce compensation mechanisms to mitigate these limitations.
One potential approach involves a critic mechanism based on multi-agent debate to cross-validate and refine the generated plan.
In addition, integrating human-in-the-loop interventions can better address challenging task scenarios and improve the overall effectiveness of the method.

This work primarily focuses on complex conditional tasks where agents need to actively gather environmental information and dynamically adapt.
However, its applicability is limited in tasks involving strong coupling and strict sequential constraints (e.g., \textit{cooking a recipe}), as well as in open-ended environments characterized by fuzzy conditions and continuous states.

\section{Ethics Statement}
We ensure our study aligns with the Code of Ethics. 
However, as an LLM-based application, it may be exploited by malicious individuals or affected by the hallucinations introduced by the specific LLM, which may generate misleading scenarios.
We urge users to apply our study responsibly.

\section{Acknowledgements}
This work was supported by the Shanghai Municipal Special Program for Basic Research on General Al Foundation Models (Grant No.2025SHZDZX026D04).

\bibliography{custom}

\vfill
\newpage

\appendix


\section{LogicEnvEval}
\label{appendix:EnvEval}

In this section, we provide a detailed description of the LogicEnvEval benchmark, which comprises 50 long-horizon household tasks featuring complex conditional logic. 
To construct this dataset, we first utilize LLMs to design diverse conditional tasks \cite{bode2024comparison} alongside their corresponding ground-truth behavior plans structured as decision trees. 
The task design principles focus on reflecting common human needs in real world. 
Subsequently, we equip every task with four distinct agent policies represented as Behavior Trees (BTs). 
These include one correct policy generated through Code-BT \cite{zhangcode} and three faulty variations. 
Based on the existing taxonomy of faulty BTs \cite{wang2025besimulator}, we construct these faulty policies by altering nodes within the correct policy. 
Finally, we conduct multiple rounds of manual verification and refinement to ensure the validity of the task data.

The four agent policies are categorized based on their adherence to physical and logical constraints. 
To be specific, the faulty policy types include Counterfactuals, Unreachable and Lackbranch.
Counterfactuals denotes that the BT violates real-world causal logic (e.g., \textit{taking an apple from a refrigerator before opening it}).
Unreachable means that the BT conforms to real-world logic but fails the task goal (e.g., \textit{merely grasping a book while the goal is to clean it}). 
The above two types consider all possible task situations but fail in one, while Lackbranch indicates that the BT omits some situations, thus failing to adapt to diverse environments.

\begin{figure*}[htbp]
    \centering
    \includegraphics[width=\textwidth]{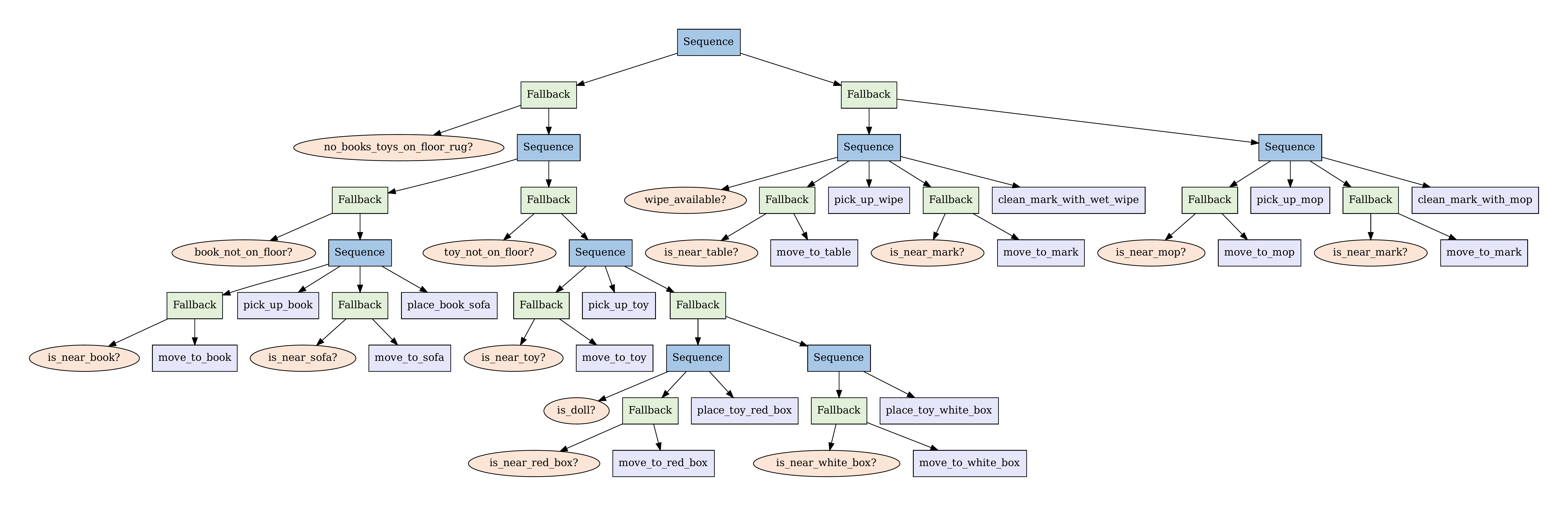}
    \caption{Correct policy (Behavior Tree) of the example \textit{Clean Living Room}.}
    \label{fig:bt_good}
\end{figure*}

\begin{figure*}[tbp]
    \centering
    \includegraphics[width=\textwidth]{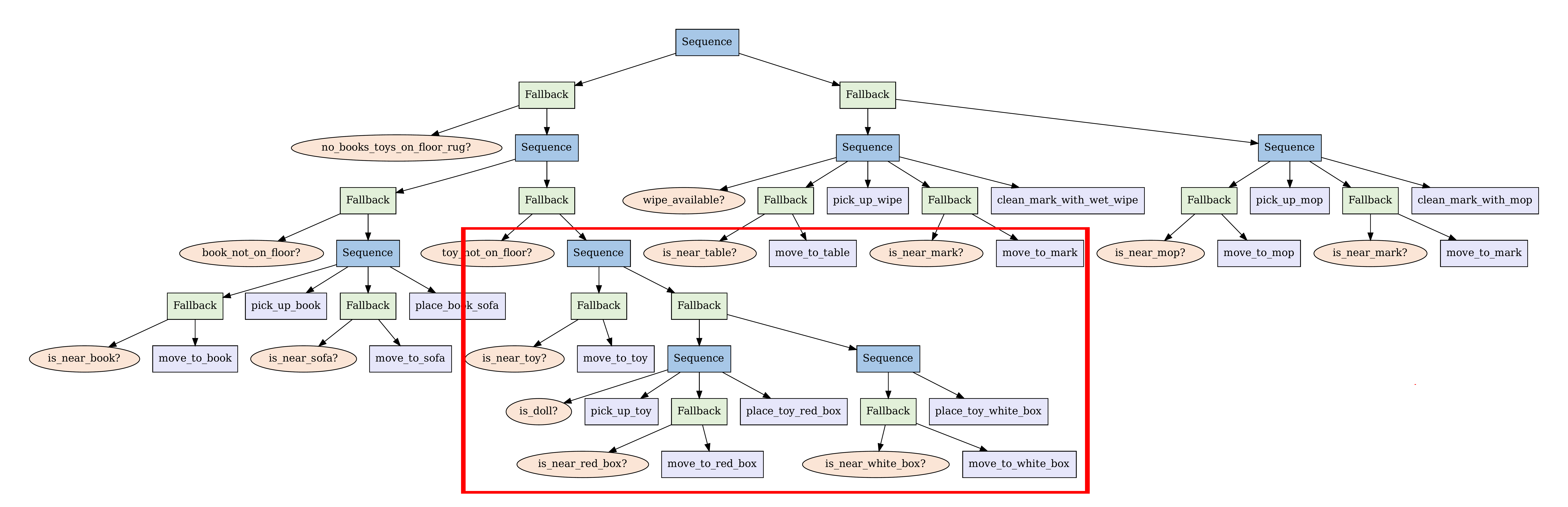}
    \caption{Faulty policy (Behavior Tree) of the example \textit{Clean Living Room}, which belongs to the Counterfactuals category. Reason: The agent does not pick up the toy (non-doll) before placing it into the white toy box.}
    \label{fig:env1}
\end{figure*}

\begin{figure*}[tbp]
    \centering
    \includegraphics[width=\textwidth]{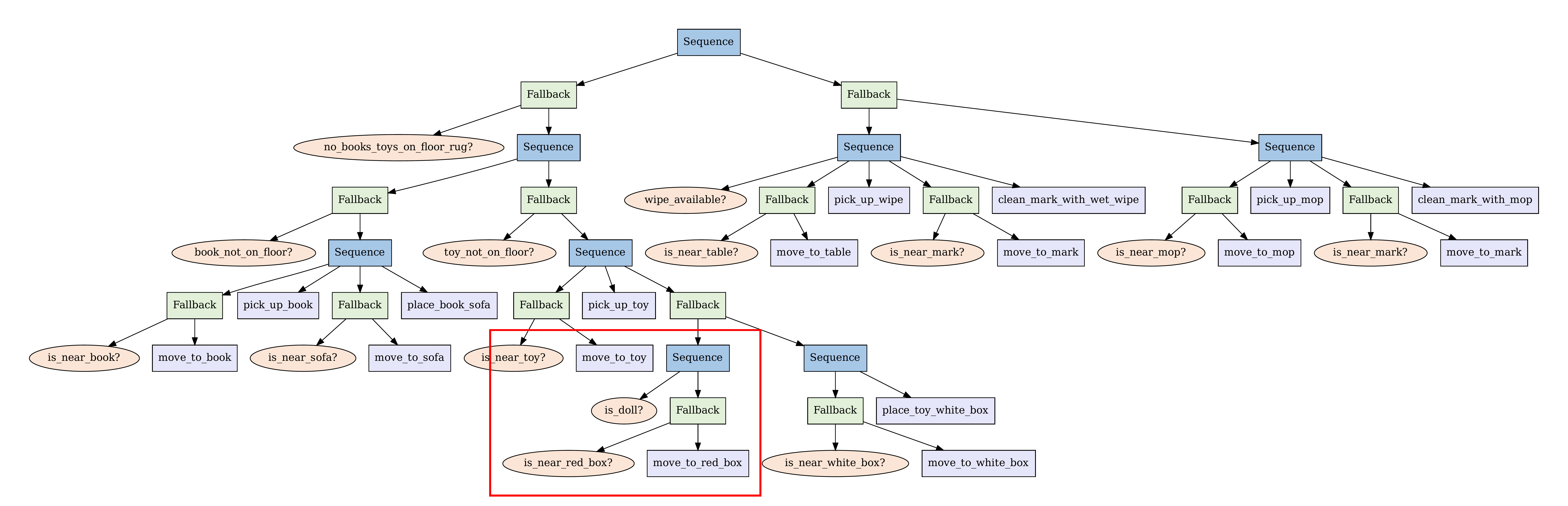}
    \caption{Faulty policy (Behavior Tree) of the example \textit{Clean Living Room}, which belongs to the Unreachable category. Reason: The agent picks up the toy (doll) but does not place it into the red toy box.}
    \label{fig:env2}
\end{figure*}

\begin{figure*}[tbp]
    \centering
    \includegraphics[width=\textwidth]{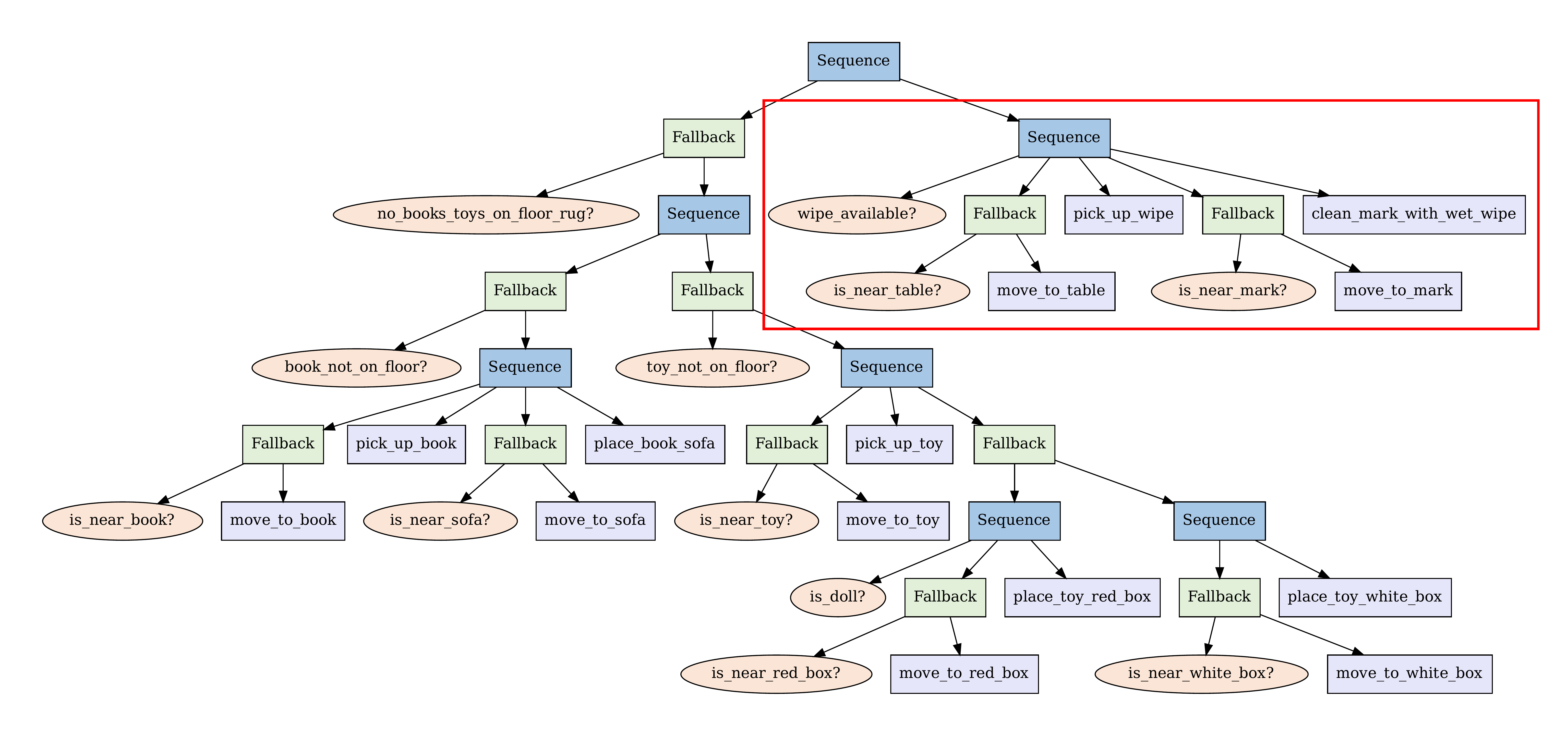}
    \caption{Faulty policy (Behavior Tree) of the example \textit{Clean Living Room}, which belongs to the Lackbranch category. Reason: The agent fails to plan for situations where wet wipes are absent.}
    \label{fig:env3}
\end{figure*}

To illustrate the structure and data format of the benchmark, we present a task example \textit{Clean Living Room}, as shown in Table \ref{tab:enveval}.
The correct agent policy is illustrated in Figure \ref{fig:bt_good}, while the three faulty variations are visualized in Figures \ref{fig:env1}, \ref{fig:env2} and \ref{fig:env3}, respectively.


\section{Baseline Methods}
\label{appendix:Baseline}

Currently, there is no specific framework designed for generating logically diverse test environments for embodied agents.
Thus, we compare LogicEnvGen with two environment generation methods, Holodeck \cite{yang2024holodeck} and I-Design \cite{I-Design}, and two scenario-agnostic methods including Chain of Thought (CoT) \cite{wei2022chain} and IFG \cite{ahmed2025intent}.
Specifically, for the CoT and IFG baselines, we generate text-based test cases and then instantiate them in AI2-THOR \cite{kolve2017ai2}, aligning with the implementation of our method.
To ensure fairness, all methods use the same problem setup, few-shot examples, and asset library.

\begin{itemize}

    \item \textbf{CoT:} We prompt LLMs to generate explicit reasoning steps and subsequently design multiple test environments based on the embodied task description to cover all potential task situations.
    
    \item \textbf{IFG:} IFG is a two-stage method that first samples intents and then generates responses to promote semantic diversity of LLM output. 
    Following the official settings \cite{ahmed2025intent}, we set the LLM temperature to 1.1 during the intent stage to generate diverse logical trajectories. 
    Subsequently, we lower the temperature to 0.5 during the final generation stage to synthesize test environments conditioned on the embodied task description and the logical trajectories of the first stage.
    
    \item \textbf{Holodeck:} Holodeck is a language-guided system capable of generating interactive 3D environments from textual descriptions. 
    We provide Holodeck embodied task descriptions to generate 3D environments.

    \item \textbf{I-Design:} I-Design leverages multi-agent collaboration to transform natural language descriptions into plausible 3D interior designs. 
    We provide I-Design with embodied task descriptions to construct simulated environments. 
    Since I-Design is constrained to single-room generation and lacks the capability to design doors or windows, we adopt a generate-and-stitch strategy to build multi-room environments, enabling a unified evaluation with other methods.
    Specifically, for multi-room tasks, we invoke I-Design multiple times to generate individual rooms independently and then stitch them together in Blender.
    
\end{itemize}


\section{More Quantitative Results}

In the Behavior Plan Derivation phase, we leverage LLM to analyze task execution logic and generate decision-tree-structured behavior plans. 
This process consists of three steps: Task decomposition, Uncertain factor identification, and Behavior plan generation. 
In this section, we present a detailed quantitative analysis of the LLM stability and accuracy across these three steps. 
Specifically, we conduct five repeated experiments on the 25 tasks of LogicEnvEval with the same setup: set the LLM temperature to 0, top-p to 1, and fix the seeds. 
The results are summarized in Table \ref{tab:analysis}.

Experimental results indicate that the LLMs are non-deterministic even with the same hyper-parameters. 
Notably, Gemini-2.5-Flash demonstrates superior stability compared to DeepSeek-V3 and Qwen2.5-72B, yielding consistent responses in every run. 
Through in-depth analysis, we find that LLMs are prone to several issues: incomplete task decomposition, missing uncertain factors, and incorrect uncertain factors.
Specifically, the LLMs occasionally group mutually independent subtasks into a single subtask, resulting in the logically redundant simulated environments. 
Furthermore, when identifying uncertain environment factors, the LLMs may overlook specific factors, leading to missed branches; alternatively, they may consider incorrect factors (e.g., \textit{the existence or functionality of an object}), which undermine the validity of simulated environments.

\begin{table}[tbp]
    \centering
    \fontsize{8.1}{10}\selectfont
    \setlength{\tabcolsep}{3.5pt}

    \begin{tabular*}{\linewidth}{l ccc}
    
        \toprule
            \textbf{Model}      & \textbf{Acc\_TD(\%)}    & \textbf{Acc\_UFI(\%)}    & \textbf{Acc\_BPG(\%)} \\
        \midrule
            DeepSeek-V3       & $96.00 \pm 0.00$     
                              & $94.56 \pm 0.36$    
                              & $95.04 \pm 0.36$ \\  
        \midrule
            Gemini-2.5-Flash  & $97.33 \pm 0.00$     
                              & $94.00 \pm 0.00$    
                              & $94.00 \pm 0.00$ \\
        \midrule
            Qwen2.5-72B       & $89.47 \pm 0.73$     
                              & $87.07 \pm 1.01$    
                              & $87.07 \pm 1.01$ \\
        \bottomrule
        
    \end{tabular*}

    \caption{Quantitative result of LLM stability and accuracy across the three steps of Behavior Plan Derivation. ``Acc\_TD'', ``Acc\_UFI'', and ``Acc\_BPG'' denote the accuracy of Task Decomposition, Uncertain Factor Identification, and Behavior Plan Generation, respectively.}
    \label{tab:analysis} 
\end{table}

\newcommand{\q}{\hspace*{1.5em}} 
\begin{table*}[htbp]
    \centering
    \fontsize{10}{12}\selectfont
    \setlength{\tabcolsep}{1pt}
    
    \begin{tabular}{m{0.25\textwidth} m{0.75\textwidth}}
        \toprule
            \textbf{Task name} & Clean Living Room \\ 
        \midrule
            \textbf{Task description} & You are in the living room. There are a sofa and a table with a wet mop leaning against it. There may be a book or a toy on the floor. The room has two toy boxes - a red one for dolls and a white one for other toys. There is a dirty mark on the floor, and you may find a pack of wet wipes on the table. \textbf{Your task:} Please check the living room floor, if there are items on floor, put them where they belong (e.g., book should be placed on the sofa, toy should be placed in the toy box). Clean the mark, preferably with the wet wipes on the table (if available). If there aren't any wipes, you can use the wet mop instead. \\
        \midrule
            \raisebox{0.5ex}{\makecell[l]{\textbf{Decision-tree-structured} \\ \textbf{behavior plan}}} & 
            \begin{tabular}[b]{@{}l@{}}
                [ \\
                \q\{ \\
                \q\q "There is a toy on the floor?": \{ \\
                \q\q\q "YES": \{ \\
                \q\q\q\q "What is the type of the toy on the floor?": \{ \\
                \q\q\q\q\q "doll": "Place the toy in the red box.", \\
                \q\q\q\q\q "other types": "Place the toy in the white box." \\
                \q\q\q\q \}, \\
                \q\q\q "NO": "Do nothing." \\
                \q\q \} \\
                \q\}, \\
                \q\{ \\
                \q\q "There is a book on the floor?": \{ \\
                \q\q\q "YES": "Place the book on the sofa.", \\
                \q\q\q "NO": "Do nothing." \\
                \q\q \} \\
                \q\}, \\
                \q\{ \\
                \q\q "There is a wet wipe on the table?": \{ \\
                \q\q\q "YES": "Clean stain with the wet wipe.", \\
                \q\q\q "NO": "Clean stain with the wet mop." \\
                \q\q \} \\
                \q\} \\
                ] \\
            \end{tabular} \\
        \bottomrule
    \end{tabular}

    \caption{Task description and decision-tree-structured behavior plan of the example \textit{Clean Living Room}.}
    \label{tab:enveval}
\end{table*}

\section{Generation Efficiency}
\label{appendix:VisualEvaluation}

Excluding the rendering process, our pipeline requires an average of 2.5 minutes per scene on a single RTX 4090 GPU.
This duration comprises behavior plan derivation (approximately 25 seconds), floor plan design (approximately 13 seconds), environment object selection (approximately 45 seconds), and constraint-based object arrangement (approximately 70 seconds). 
For comparison, the generation processes of Holodeck \cite{yang2024holodeck} and I-Design \cite{I-Design} require approximately 2.3 and 3.7 minutes per scene, respectively.



\begin{figure*}[htbp]
    \centering
    \includegraphics[width=\textwidth]{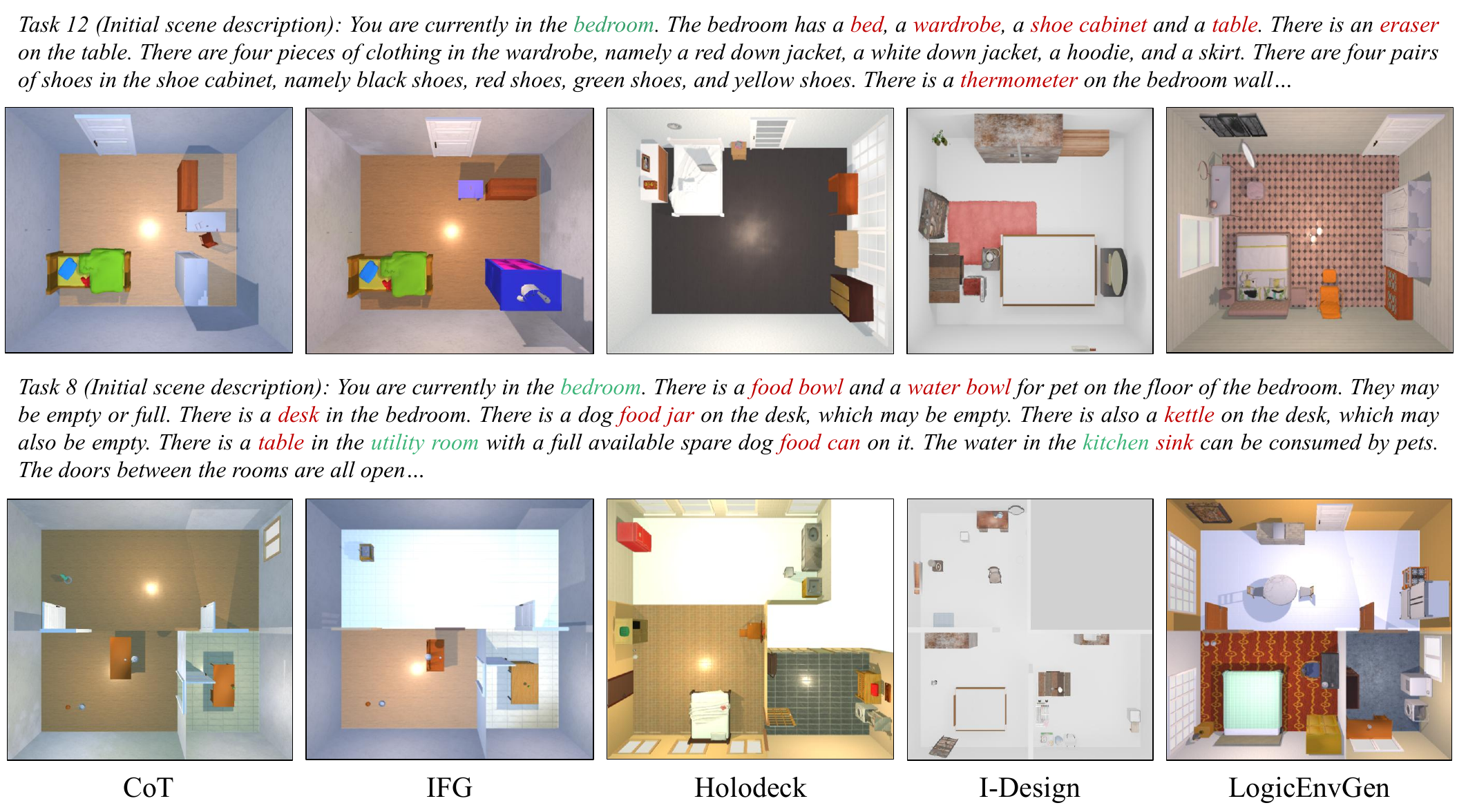}
    \caption{Qualitative comparison on two LogicEnvEval tasks. While baseline methods suffer from physical implausibilities (e.g., floating objects), irrational layout or missing task-relevant items, LogicEnvGen generates environments with superior visual fidelity, layout coherence, and strict task alignment.}
    \label{fig:Visual_examples}
\end{figure*}

\newpage

\section{Visual Evaluation}
\label{appendix:VisualEvaluation}

To better understand the potential trade-off between logical diversity and visual realism, we conduct the qualitative and quantitative evaluations to compare the visual quality of the environments generated by LogicEnvGen and baselines.
Four baselines and LogicEnvGen utilize the same Objaverse assets to ensure a fair comparison. 
For 25 tasks sampled from LogicEnvEval, we randomly select one generated environment per method.
We group the environments of the same task from five methods, resulting in 25 groups of environments for visual evaluation.

\subsection{Qualitative Experiment}
As shown in Figure \ref{fig:Visual_examples}, we present the visual results of environments generated by five methods across two LogicEnvEval tasks. 
Compared to the four baselines, LogicEnvGen demonstrates significant advantages in visual fidelity, layout coherence, and task alignment. 

Specifically, while CoT generates necessary task-relevant objects, it suffers from severe violations of physical constraints (e.g., \textit{floating kettles in Task 8}) and irrational object placements. 
Similarly, IFG incorporates necessary objects but lacks object diversity and exhibits incoherent spatial arrangements (e.g., \textit{large furniture not being placed flush against the walls}). 
Holodeck excels in object richness and layout consistency but struggles to satisfy task constraints, often omitting critical task-relevant items (e.g., \textit{food cans in Task 8}). 
While I-Design demonstrates reasonable layout coordination, it frequently exhibits visual distortions (e.g., \textit{bookshelf in Task 12}) and lacks task-relevant objects.
In contrast, LogicEnvGen achieves visual realism comparable to Holodeck while ensuring adherence to task descriptions.

\subsection{Quantitative Experiment}

\begin{figure*}[htbp]
    \centering
    \includegraphics[width=\textwidth]{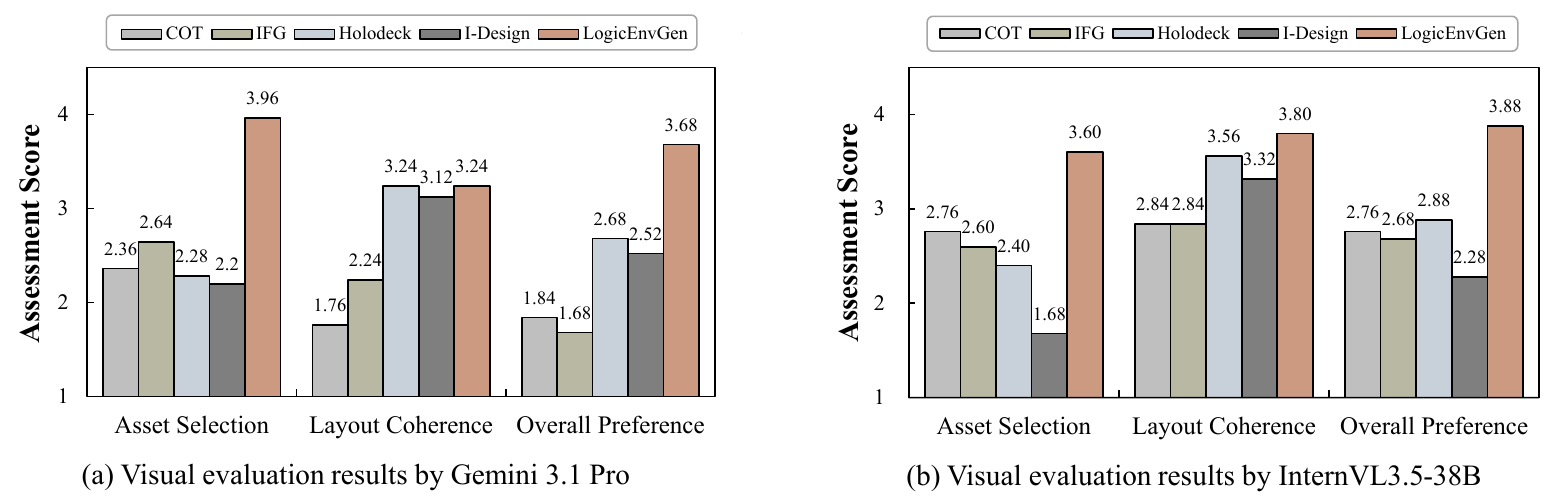}
    \caption{VLM evaluation of LogicEnvGen and four baselines. The results from both Gemini 3.1 Pro and InternVL3.5-38B consistently show LogicEnvGen outperforms four baselines across three criteria.}
    \label{fig:VLM_evaluation}
\end{figure*}

\subsubsection{Setup}
In this section, we conduct automatic evaluation and human user evaluation to assess the quality of generated 3D environments.
The evaluation is primarily based on three criteria:
(1) \textbf{Asset Selection}: Which environment selects 3D assets that are more accurate and faithful to the task description and the environment type?
(2) \textbf{Layout Coherence}: Which environment arranges 3D assets in a more realistic and logically consistent manner?
(3) \textbf{Overall Preference}: Given the environment type and task description, which environment is preferred overall in terms of visual quality (realism and richness) and utility?
Notably, Layout Coherence assesses pure visual quality, whereas the other two criteria incorporate task alignment.

\subsubsection{Automatic Evaluation}

To enable automatic quantitative evaluations, we leverage Vision Language Models (VLMs) as evaluators to provide an objective assessment for the generated environments.
We employ two well-known VLMs in the field of closed source and open source, including Gemini 3.1 Pro \cite{gemini-3-pro} and InternVL3.5-38B \cite{wang2025internvl3_5}.
To mitigate positional bias, each group of environments is presented to the evaluation model as five shuffled top-down view images. 
The evaluation model is asked to score the environments on a scale of 1 to 5 (with higher scores indicating better quality) based on the three criteria.

Figure \ref{fig:VLM_evaluation} demonstrates that LogicEnvGen achieves substantially higher scores in both Asset Selection and Overall Preference across two VLM evaluations. 
Even for the Layout Coherence dimension, in which baselines such as Holodeck exhibit competitive results, LogicEnvGen maintains comparable or superior performance. 
These results indicate that LogicEnvGen generates environments that are more realistic and better aligned with the target tasks.

\begin{figure}[tp]
    \centering
    \includegraphics[width=\linewidth]{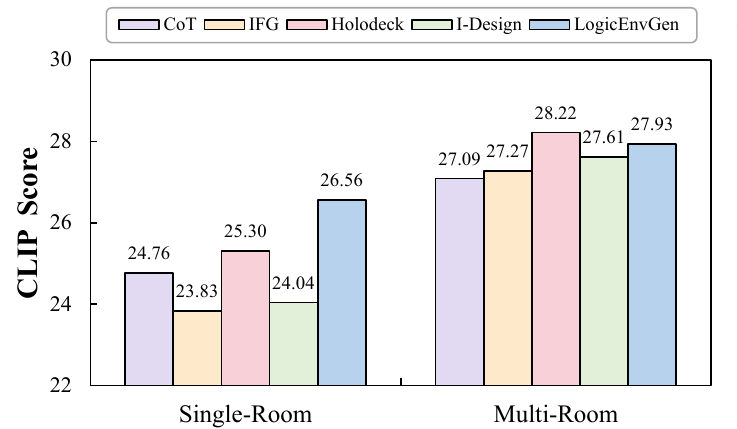}
    \caption{CLIP Score comparison over two scenario categories.}
    \label{fig:CLIP}
\end{figure}

In addition to VLM evaluation, we utilize CLIP Score\footnote{Here, we use OpenAI ViT-L/14 model. We use cosine similarity times 100 as the CLIP Score.} \cite{hessel2021clipscore} to assess the visual consistency between the top-down view of the environment and its designated environment type, formulated as the text prompt ``a top-down view of [environment type]'' (e.g., \textit{``a top-down view of living room''}).
We categorize the tasks in LogicEnvEval into two categories based on task scenario: Single-Room and Multi-Room.
Figure \ref{fig:CLIP} demonstrates that LogicEnvGen outperforms all four baselines in the Single-Room category. 
In the Multi-Room category, it achieves performance comparable to Holodeck while surpassing CoT, IFG and I-Design. 
These results validate LogicEnvGen's capability to generate environments that are visually consistent with the designated environment types.

\begin{figure}[tp]
    \centering
    \includegraphics[width=\linewidth]{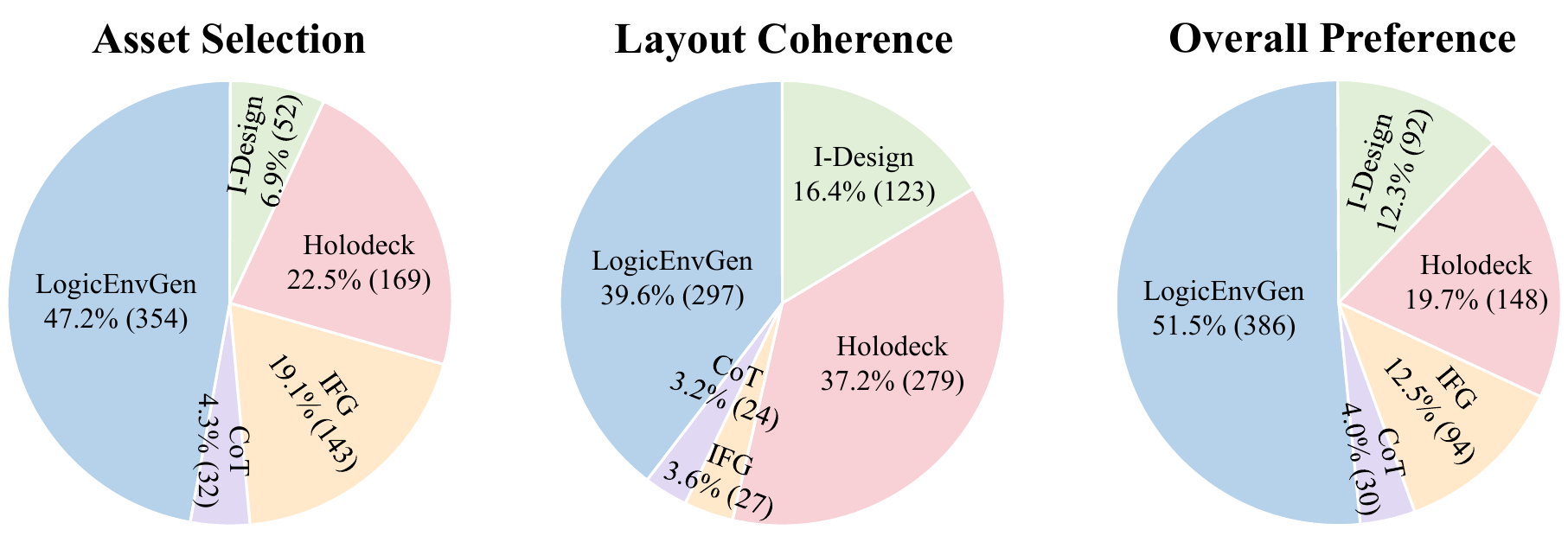}
    \caption{Comparative human evaluation of LogicEnvGen and four baselines across three criteria. The pie charts show the distribution of annotator preferences, showing both the percentage and the actual number of annotations favoring each method.}
    \label{fig:Human_evaluation}
\end{figure}

\begin{table*}[htbp]
    \centering
    \fontsize{9.1}{11}\selectfont
    \setlength{\tabcolsep}{2.1pt}

    \begin{tabular*}{0.95\textwidth}{@{\extracolsep{\fill}} l c c c c c c c}
        \toprule
            \multirow{2}{*}{\textbf{Method}} & \multirow{2}{*}{\textbf{PhyPR(\%)}} & \multirow{2}{*}{\textbf{LogCov(\%)}} & \multirow{2}{*}{\textbf{SceVR(\%)}} & \multicolumn{4}{c}{\textbf{Fault Detection Rate (\%)}} \\
            
            \cmidrule(lr){5-8}
        
            & & & & \textbf{CFactuals} & \textbf{Unreach} & \textbf{LBranch} & \textbf{Avg} \\

            \midrule
            Holodeck       & 100.00 & 37.07 & 74.00 & 24.00 & 22.00 & 28.00 & 24.67 \\
            Holodeck+Logic & 100.00 & 87.92 & 90.26 & 76.00 & 84.00 & 78.00 & 79.33 \\
            \textbf{LogicEnvGen}    & 100.00 & \textbf{98.97} & \textbf{92.40} & \textbf{94.00} & \textbf{90.00} & \textbf{94.00} & \textbf{92.67} \\
            \bottomrule
    \end{tabular*}

    \caption{\label{tab:attribution_ablation}
    Attribution ablation results on the LogicEnvEval benchmark.}
\end{table*}

\subsubsection{Human Evaluation}

We conduct human evaluations involving 30 graduate students.
This study collects human preference scores to compare LogicEnvGen with four baselines.
For each group of environments, we display five randomly shuffled top-down view images of the environments from the five methods.
Annotators are asked to select the optimal environment based on the three predefined criteria.

Figure \ref{fig:Human_evaluation} illustrates a clear preference for LogicEnvGen over the four baselines in the human evaluation. 
Specifically, annotators favor LogicEnvGen in Asset Selection (47.2\%), Layout Coherence (39.6\%), and show a significant preference in Overall Preference (51.5\%). 
The human evaluation experiment agrees with the automatic evaluation.


\section{Attribution Ablation Study}
\label{sec:attribution_ablation}

To strictly isolate the contribution of the logical trajectory generation from the environment construction module, we conduct an attribution ablation study.
We feed the logical trajectories generated by our framework explicitly as text prompts into the Holodeck baseline, denoted as Holodeck+Logic.
Table \ref{tab:attribution_ablation} presents the experimental results using the DeepSeek-v3.2 model on the LogicEnvEval benchmark.

The results show that explicitly providing logical trajectories significantly raises the LogCov of Holodeck from 37.07\% to 87.92\%.
However, a crucial gap remains in SceVR and FauDR.
While Holodeck successfully generates plausible rooms, it frequently fails to adhere to the dense task-specific constraints specified in the logical trajectories and task descriptions.
For instance, Holodeck fails to initialize a kabob on the plate, which is required by the task description.
Because its pipeline biases toward general visual richness rather than strict task alignment, these violations render the generated environments invalid.
Thus, its SceVR and average FauDR fall short of our method.


\section{Manual Spot Check of Environment Validity}

In the experiments, we evaluate SceVR by employing a behavior simulator to simulate the correct behavior tree within the generated environments.
An environment is classified as valid if the correct behavior tree is evaluated as fault-free by the simulator.
This automated verification process essentially functions as a validity checker.
To validate its reliability, we conduct a manual blind spot check.
We randomly sample 110 simulated environments generated across all baselines and LogicEnvGen.
Human annotators judge the validity of each environment based on the predefined SceVR criteria, which include verifying the presence of task-related objects and the satisfaction of explicit task constraints.
To prevent confirmation bias, we keep this evaluation strictly blind.
The annotators are unaware of both the generating method and the automated verification result.
The evaluation results are presented in Table \ref{tab:validity_checker}.

The results show a 99.09\% agreement rate between the human annotators and the automated validity checker.
This high consistency confirms that the checker is highly reliable.
It does not skew the FauDR metric by letting invalid environments slip through or falsely rejecting valid ones.
We carefully analyze the single discrepancy where the checker passes the environment as valid while the human initially flags it as invalid.
This case occurs because the environment partially aligns with the task description but perfectly satisfies the specific constraints of the associated logical trajectory.
Thus, the correct behavior tree successfully executes and is evaluated as fault-free.
From a software testing perspective, the environment is operationally sound for evaluating that specific logic branch.
Therefore, classifying this environment as valid is technically accurate and preserves the fairness of the FauDR evaluation when testing faulty behavior trees.

\begin{table}[h]
    \centering
    \fontsize{8.6}{11}\selectfont
    \setlength{\tabcolsep}{2.1pt}
    
    \begin{tabular}{lcc}
        \toprule
        & Checker: Valid & Checker: Invalid \\
        \midrule
        Human Evaluator: Valid & 89 & 0 \\
        Human Evaluator: Invalid & 1 & 20 \\
        \bottomrule
    \end{tabular}
    \caption{Confusion matrix of the validity checker against human evaluation.}
    \label{tab:validity_checker}
\end{table}



\section{Case Study}
\label{appendix:CaseStudy}
\begin{figure*}[tbp]
    \centering
    \includegraphics[width=0.7\textwidth]{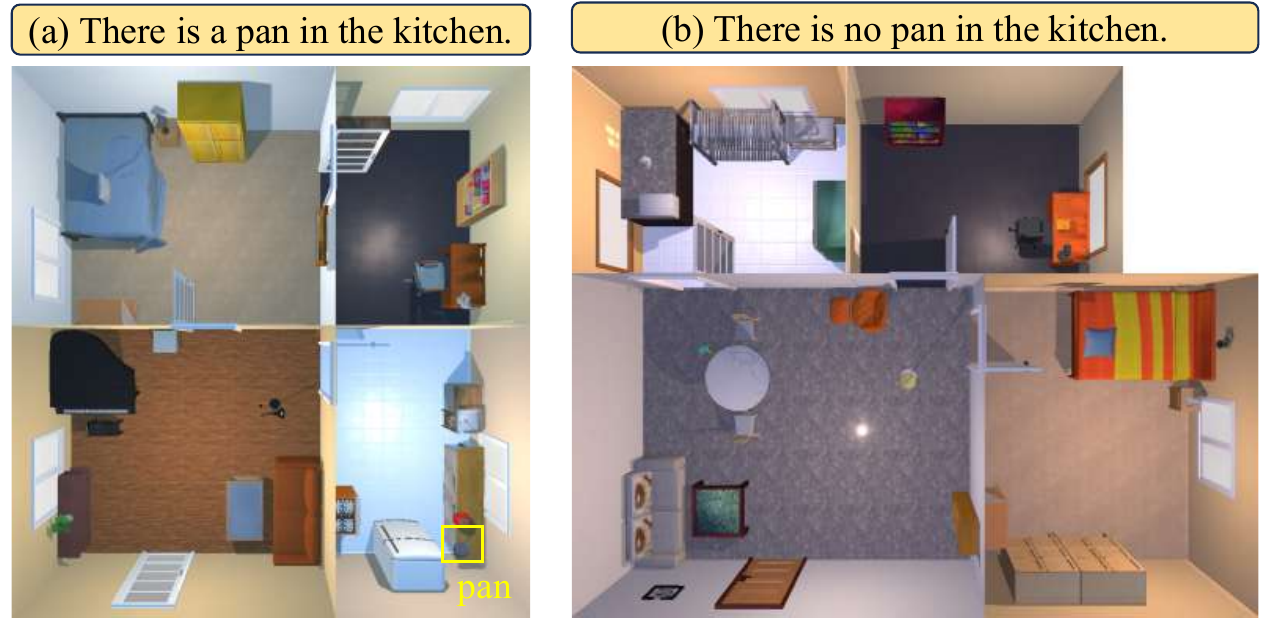}
    \caption{Case 1. Agent task: Bring me a desk lamp from study if there is a pan in the kitchen. Otherwise, bring me a pillow from the bedroom.}
    \label{fig:case1}
\end{figure*}

\begin{figure*}[tbp]
    \centering
    \includegraphics[width=0.69\textwidth]{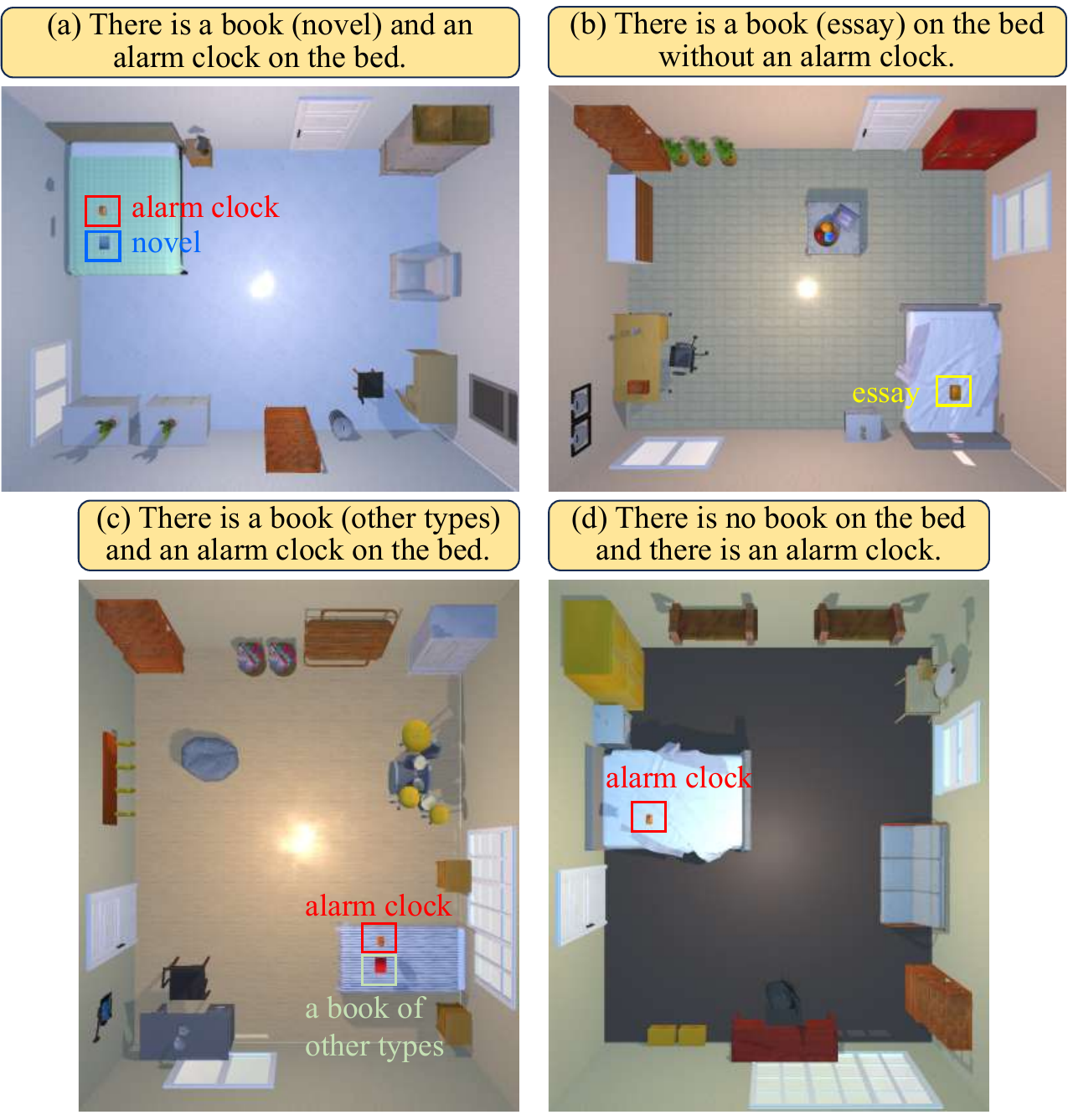}
    \caption{Case 2. Agent task: The bedroom has a bed, possibly with a book or clock on it, along with a bookshelf and a desk. The bookshelf's three levels hold novels, essays, and other books respectively. Clean items on the bed and put them into their designated storage areas (e.g., book on the shelf, phone on the desk). }
    \label{fig:case2}
\end{figure*}

\begin{figure*}[tbp]
    \centering
    \includegraphics[width=0.7\textwidth]{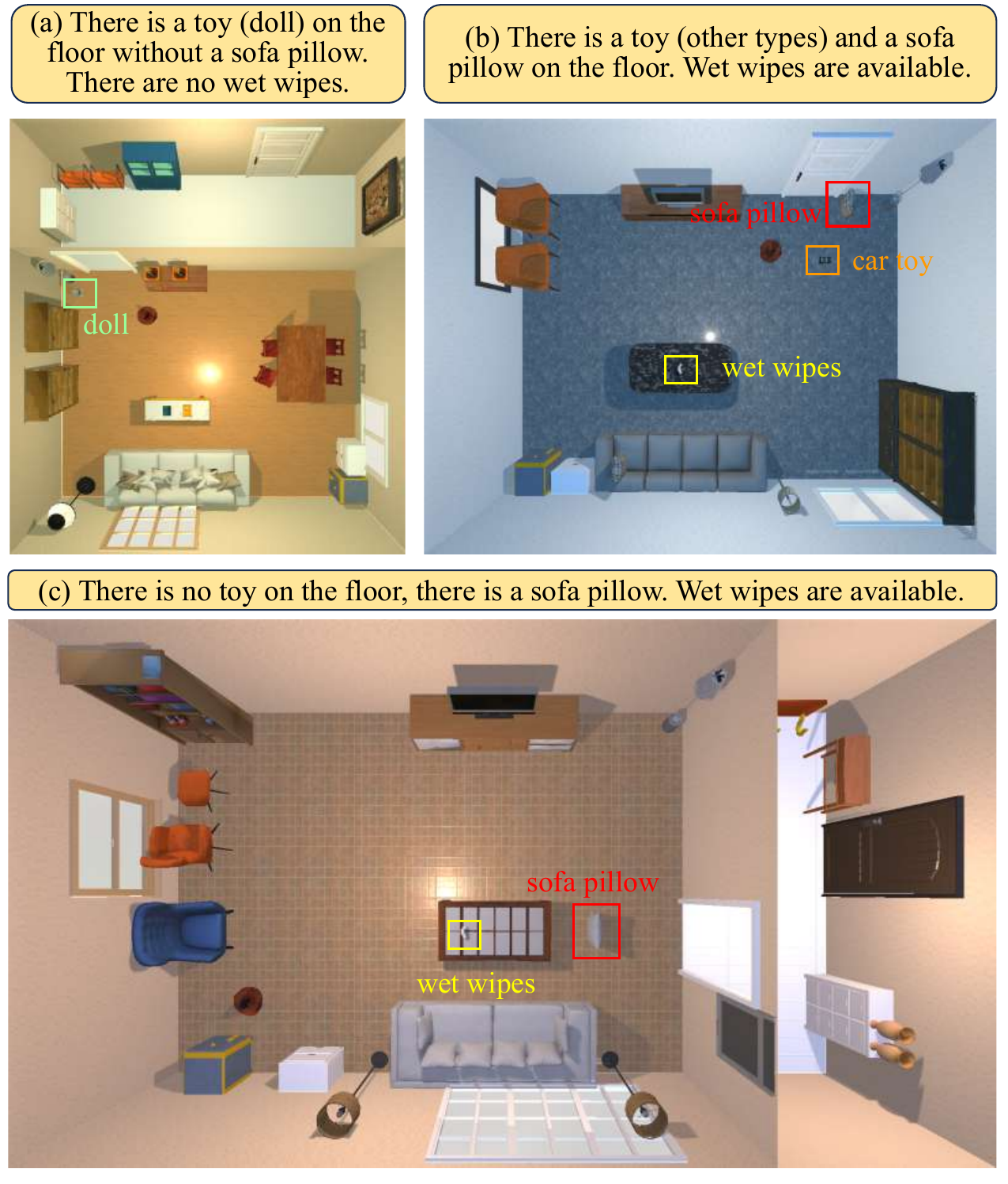}
    \caption{Case 3. Agent task: The living room has a sofa, a tea table, and two toy boxes—the blue one for dolls and the white one for other toys. The floor is stained and there might be a sofa pillow or toy on it. Check the floor and put misplaced items (like a pillow or toy) in their proper places. Clean the stain, preferably with the wet wipes on the table (if available), otherwise use the wet mop by the wall.}
    \label{fig:case3}
\end{figure*}

In Figures \ref{fig:case1}, \ref{fig:case2} and \ref{fig:case3}, we show the simulation test cases generated by LogicEnvGen for three embodied tasks.

\end{document}